\documentclass[]{article}
\usepackage{proceed2e}
\usepackage{times}
\usepackage{natbib}
\usepackage[ruled,vlined]{algorithm2e}
\usepackage{amsmath}
\usepackage{amssymb}
\usepackage{graphicx}
\usepackage{enumerate}

\title{Parallel Gaussian Process Regression with\\ Low-Rank Covariance Matrix Approximations} 
\author{Jie Chen$^{\dag}$, Nannan Cao$^{\dag}$, Kian Hsiang Low$^{\dag}$, Ruofei Ouyang$^{\dag}$, Colin Keng-Yan Tan$^{\dag}$, and Patrick Jaillet$^{\S}$\\
  Department of Computer Science, National University of Singapore, Republic of Singapore$^{\dag}$\\
  Department of Electrical Engineering and Computer Science,
  Massachusetts Institute of Technology, USA$^{\S}$
}

\newcounter{sol} 
\begin{document} 
\newcommand{\set}[1]{{\mathcal #1}}
\newcommand{\sset}[1]{{|\mathcal #1|}}
\newcommand{\bigo}[1]{{\mathcal{O}\hspace{-1mm} \left( #1 \right)}}
\newcommand{\squishlisttwo}{
 \begin{list}{$\bullet$}
  { \setlength{\itemsep}{0pt}
     \setlength{\parsep}{0pt}
    \setlength{\topsep}{0pt}
    \setlength{\partopsep}{0pt}
    \setlength{\leftmargin}{1em}
    \setlength{\labelwidth}{1.5em}
    \setlength{\labelsep}{0.5em} } }

\newcommand{\squishend}{
  \end{list}  }

\newtheorem{defi}{Definition}
\newtheorem{thm}{Theorem}
 
\maketitle 
 
\begin{abstract}\vspace{-4.4mm}
\emph{Gaussian processes} (GP) are Bayesian non-parametric models that
are widely used for probabilistic regression. 
Unfortunately, it cannot scale well with large data nor perform real-time predictions due to its cubic time cost in the data size. 
This paper presents two parallel GP regression methods that exploit low-rank covariance matrix approximations for distributing the computational load among parallel machines to achieve time efficiency and scalability.
We theoretically guarantee the predictive performances of our proposed parallel GPs to be equivalent to that of some centralized approximate GP regression methods: The computation of their centralized counterparts can be distributed among parallel machines, hence achieving greater time efficiency and scalability. 
We analytically compare the properties of our parallel GPs such as time, space, and communication complexity.
Empirical evaluation on two real-world datasets in a cluster of $20$ computing nodes shows that our parallel GPs are significantly more time-efficient and scalable than their centralized counterparts and exact/full GP while achieving predictive performances comparable to full GP.
\vspace{-4.5mm}
\end{abstract}

\section{Introduction}
\label{sec:intro}
\vspace{-3mm}
\emph{Gaussian processes} (GP) are Bayesian non-parametric models for performing nonlinear regression, which offer an important advantage of providing fully probabilistic predictive distributions with formal measures of the uncertainty of the predictions.
The key limitation hindering the practical use of GP for large data is the high computational cost: It incurs cubic time and quadratic memory in the size of the data.
To reduce the computational cost, two classes of approximate GP regression methods have been proposed: (a) Low-rank covariance matrix approximation methods \citep{candela05,snelson06,Williams00} are especially suitable for modeling smoothly-varying functions with high correlation (i.e., long length-scales) and they utilize all the data for predictions like the exact/full GP; and (b) localized regression methods (e.g., local GPs \citep{Das10,Choudhury02,Park11} and compactly supported covariance functions \citep{Furrer06}) are capable of modeling highly-varying functions with low correlation (i.e., short length-scales) but they use only local data for predictions, hence predicting poorly in input regions with sparse data.
Recent approximate GP regression methods of \cite{snelson07} and \cite{Vanhatalo08} have attempted to combine the best of both worlds.

Despite these various efforts to scale up GP, it remains computationally impractical for performing real-time predictions necessary in many time-critical applications and decision support systems (e.g., ocean sensing \citep{LowAAMAS13,LowSPIE09,LowICRA07,LowAAMAS11,LowAAMAS12,LowAeroconf10}, traffic monitoring \citep{chen12,LowIAT12}, geographical information systems) that need to process and analyze huge quantities of data collected over short time durations (e.g., in astronomy, internet traffic, meteorology, surveillance). To resolve this, the work in this paper considers exploiting clusters of parallel machines to achieve efficient and scalable predictions in real time. Such an idea of scaling up machine learning techniques (e.g., clustering, support vector machines, graphical models) has recently attracted widespread interest in the machine learning community \citep{Bekkerman11}. For the case of Gaussian process regression, the local GPs method \citep{Das10,Choudhury02} appears most straightforward to be ``embarrassingly'' parallelized but they suffer from discontinuities in predictions on the boundaries of different local GPs. The work of \cite{Park11} rectifies this problem by  imposing continuity constraints along the boundaries in a centralized manner.
But, its use is restricted strictly to data with $1$- and $2$-dimensional input features.

This paper presents two parallel GP regression methods (Sections~\ref{sect:pgps} and~\ref{sect:pgpi})
that, in particular, exploit low-rank covariance matrix approximations for distributing the computational load among parallel machines to achieve time efficiency and scalability. 
Different from the above-mentioned parallel local GPs method, our proposed parallel GPs do not suffer from boundary effects, work with multi-dimensional input features, and exploit all the data for predictions but do not incur the cubic time cost of the full/exact GP.
The specific contributions of our work include:\vspace{-1mm}
\squishlisttwo
\item Theoretically guaranteeing the predictive performances of our parallel GPs (i.e., \emph{parallel partially independent conditional} ($p$PIC) and \emph{parallel incomplete Cholesky factorization} ($p$ICF)-based approximations of GP regression model) 
 to be equivalent to that of some centralized approaches to approximate GP regression (Sections~\ref{sect:pgps} and~\ref{sect:pgpi}). An important practical implication of these results is that the computation of their centralized counterparts can be distributed among a cluster of parallel machines, hence achieving greater time efficiency and scalability. Furthermore, our parallel GPs 
inherit an advantage of their centralized counterparts in providing a parameter (i.e., size of support set for $p$PIC and reduced rank for $p$ICF-based GP) to be adjusted in order to trade off between predictive performance and time efficiency;
\item Analytically comparing the properties of our parallel GPs such as time, space, and communication complexity, capability of online learning, and practical implications of the structural assumptions (Section~\ref{sect:compare});
\item Implementing our parallel GPs using the \emph{message passing interface} (MPI) framework to run in a cluster of $20$ computing nodes and empirically evaluating their predictive performances, time efficiency, scalability, and speedups on two real-world datasets (Section~\ref{sec:exp}).\vspace{-1.2mm}
\squishend
\section{Gaussian Process Regression}
\label{sect:gpr}
\vspace{-3mm}
The \emph{Gaussian process} (GP) can be used to perform probabilistic regression as follows: Let $\set{X}$ be a set representing the input domain such that each input $x\in \set{X}$ denotes a $d$-dimensional feature vector and is associated with a realized output value $y_x$ (random output variable $Y_x$) if it is observed (unobserved). 
Let $\{Y_x\}_{x \in \set{X}}$ denote a GP, that is, every finite subset of $\{Y_x\}_{x \in \set{X}}$ follows a multivariate Gaussian distribution \citep{rasmussen06}. Then, the GP is fully specified by its \emph{prior} mean $\mu_{x} \triangleq \mathbb{E}[Y_x]$ and covariance $\sigma_{x x'} \triangleq \mbox{cov}[Y_x, Y_{x'}]$ for all $x, x' \in \set{X}$.

Given that a column vector $y_\set{D}$ of realized outputs is observed for some set $\set{D}\subset\set{X}$ of inputs,
the GP can exploit this data $(\set{D},y_{\set{D}})$ to provide predictions of the unobserved outputs
for any set $\set{U}\subseteq \set{X}\setminus \set{D}$ of inputs and their corresponding predictive uncertainties using the following Gaussian \emph{posterior} mean vector and covariance matrix, respectively:\vspace{-3mm}
\begin{equation}
\displaystyle\mu_{\set{U}|\set{D}}\triangleq \mu_\set{U} + \Sigma_{\set{U}\set{D}}\Sigma_{\set{D}\set{D}}^{-1}( y_\set{D} - \mu_\set{D} )
\label{pmu}\vspace{-0mm}
\end{equation}
\begin{equation}
\displaystyle\Sigma_{\set{U}\set{U}|\set{D}}\triangleq\Sigma_{\set{U}\set{U}}-\Sigma_{\set{U}\set{D}}\Sigma_{\set{D}\set{D}}^{-1}\Sigma_{\set{D}\set{U}}
\label{pcov}\vspace{-0mm}
\end{equation}
where ${\mu}_\set{U}$ (${\mu}_\set{D}$) is a column vector with mean components $\mu_{x}$ for all $x\in \set{U}$ ($x\in \set{D}$),
$\Sigma_{\set{U}\set{D}}$ ($\Sigma_{\set{D}\set{D}}$) is a covariance matrix with covariance components $\sigma_{xx'}$ for all $x\in \set{U}, x'\in \set{D}$ ($x, x'\in \set{D}$), and $\Sigma_{\set{D}\set{U}}$ is the transpose of $\Sigma_{\set{U}\set{D}}$. 
The uncertainty of predicting the unobserved outputs can be measured using the trace of $\Sigma_{\set{U}\set{U}|\set{D}}$ (\ref{pcov}) (i.e., sum of posterior variances $\Sigma_{xx|\set{D}}$ over all $x\in\set{U}$), which is independent of the realized outputs $y_\set{D}$.\vspace{-1mm}
%
\section{Parallel Gaussian Process Regression using Support Set}
\label{sect:pgps}\vspace{-2mm}
The centralized approach to exact/full GP regression described in Section~\ref{sect:gpr}, which we call the \emph{full Gaussian process} (FGP), unfortunately cannot scale well and be performed in real time due to its cubic time complexity in the size $|\set{D}|$ of the data.
In this section, we will present a class of parallel Gaussian processes ($p$PITC and $p$PIC) that distributes the computational load among parallel machines to achieve efficient and scalable approximate GP regression by exploiting the notion of a support set.

The \emph{parallel partially independent training conditional} ($p$PITC) approximation of FGP model is adapted from our previous work on decentralized data fusion \citep{chen12} for sampling environmental phenomena with mobile sensors. But, the latter does not address the practical implementation issues of parallelization on a cluster of machines nor demonstrate scalability with large data.
So, we present $p$PITC here under the setting of parallel machines and then show how 
its shortcomings can be overcome by extending it to $p$PIC.
The key idea of $p$PITC is as follows: After distributing the data evenly among $M$ machines (Step $1$), each machine encapsulates its local data, based on a common prior support set $\set{S}\subset X$ where $\sset{S}\ll\sset{D}$, into a local summary that is communicated to the master\footnote{One of the $M$ machines can be assigned to be the master.} (Step $2$). The master assimilates the local summaries into a global summary (Step $3$), which is then sent back to the $M$ machines to be used for predictions distributed among them (Step $4$). 
These steps are detailed below:\vspace{1mm}

{\scshape Step 1: Distribute data among $M$ machines.}

The data $(\set{D},y_{\set{D}})$ is partitioned evenly into $M$ blocks, each of which is assigned to a machine, as defined below:
\begin{defi}[Local Data] The local data of machine $m$ is defined as a tuple $(\set{D}_m, y_{\set{D}_m})$ where $\set{D}_m\subseteq\set{D}$, $\set{D}_m \bigcap \set{D}_i = \emptyset$ and $|\set{D}_m| = |\set{D}_i| = |\set{D}|/M$ for $i\neq m$.\vspace{1mm} 
\label{def:ld}
\end{defi}
%

{\scshape Step 2: Each machine constructs and sends local summary to master.}
\begin{defi}[Local Summary] Given a common support set $\set{S}\subset\set{X}$ known to all $M$ machines and the local data $(\set{D}_m, y_{\set{D}_m})$, 
the local summary of machine $m$ is defined as a tuple $(\dot{y}_\set{S}^m, \dot{\Sigma}_{\set{S}\set{S}}^m)$ where
\begin{equation}
\dot{y}_{\set{B}}^m \triangleq \Sigma_{\set{B}\set{D}_m}\Sigma^{-1}_{\set{D}_m
\set{D}_m|\set{S}}\left(y_{\set{D}_m}-\mu_{\set{D}_m}\right)
\label{eq:lsf} \end{equation}
\begin{equation}
\dot{\Sigma}_{\set{B}\set{B}'}^m\triangleq
\Sigma_{\set{B}\set{D}_m}\Sigma_{\set{D}_m \set{D}_m|\set{S}}^{-1}\Sigma_{\set{D}_m \set{B}'}
\label{eq:lsc}
\end{equation}
such that $\Sigma_{\set{D}_m \set{D}_m|\set{S}}$ is defined in a similar manner as
(\ref{pcov}) and $\set{B}, \set{B}'\subset\set{X}$. 
\label{def:ls}
\end{defi}
\emph{Remark}. Since the local summary is independent of the outputs $y_{\set{S}}$, they need not be observed. So, the support set $\set{S}$ does not have to be a subset of $\set{D}$ and can be selected prior to data collection.
Predictive performances of $p$PITC and $p$PIC are sensitive to the selection of $\set{S}$.
An informative support set $\set{S}$ can be selected from domain $\set{X}$ using an iterative greedy active selection procedure \citep{Guestrin08,lawrence03,seeger03} prior to observing data.
For example, the differential entropy score criterion \citep{lawrence03} can be used to greedily select an input $x\in\set{X}\setminus\set{S}$ with the largest posterior variance $\Sigma_{xx|\set{S}}$ (\ref{pcov}) to be included in $\set{S}$ in each iteration.\vspace{1mm}

{\scshape Step 3: Master constructs and sends global summary to $M$ machines.}
\begin{defi}[Global Summary] Given a common support set $\set{S}\subset\set{X}$
known to all $M$ machines and the local summary $(\dot{y}_\set{S}^m, \dot{\Sigma}_{\set{S}\set{S}}^m)$ of every machine $m = 1,\ldots,M$, the global
summary is defined as a tuple $(\ddot{y}_\set{S}, \ddot{\Sigma}_{\set{S}\set{S}})$
where\vspace{-3mm}
\begin{equation}
\ddot{y}_\set{S} \triangleq\sum_{m=1}^{M}\dot{y}_{\set{S}}^m\vspace{-3mm}
\label{eq:gsf} \end{equation}
\begin{equation}
\ddot{\Sigma}_{\set{S}\set{S}} \triangleq\Sigma_{\set{S}\set{S}}+\sum_{m=1}^{M} \dot{\Sigma}_{\set{S}\set{S}}^m \ .
\vspace{-3mm}
\label{eq:gsc}
\end{equation}
\label{def:gs}
\end{defi}

{\scshape Step 4: Distribute predictions among $M$ machines.}

To predict the unobserved outputs for any set $\set{U}$ of inputs, $\set{U}$ is partitioned evenly into disjoint subsets $\set{U}_1, \ldots, \set{U}_M$ to be assigned to the respective machines $1,\ldots,M$. So, $|\set{U}_m|=|\set{U}|/M$ for $m=1,\ldots,M$.
\begin{defi}[$p$PITC]
Given a common support set $\set{S}\subset\set{X}$
known to all $M$ machines and the global summary $(\ddot{y}_\set{S}, \ddot{\Sigma}_{\set{S}\set{S}})$, each machine $m$ computes a predictive Gaussian distribution $\mathcal{N}(\widehat{\mu}_{\set{U}_m},\widehat{\Sigma}_{\set{U}_m\set{U}_m})$ of the unobserved outputs for the set $\set{U}_m$ of inputs where
\vspace{-1mm}
%
\begin{equation}
    \widehat{\mu}_{\set{U}_m} \triangleq\mu_{\set{U}_m} + \Sigma_{\set{U}_m\set{S}}\ddot{\Sigma}_{\set{S}\set{S}}^{-1}\ddot{y}_\set{S}
\label{mupitc}
\vspace{-1mm}
\end{equation}
\begin{equation}
  \widehat{\Sigma}_{\set{U}_m\set{U}_m} \triangleq\Sigma_{\set{U}_m\set{U}_m}-\Sigma_{\set{U}_m\set{S}}\left(\Sigma_{\set{S}\set{S}}^{-1}-\ddot{\Sigma}
_{\set{S}\set{S}}^{-1}\right)\Sigma_{\set{S}\set{U}_m}\ .
\label{varpitc}\vspace{-6mm}
\end{equation}
\label{def:ppic}
\end{defi}
\begin{thm} {\bf [\cite{chen12}]\;}
Let a common support set $\set{S}\subset\set{X}$ be known to all $M$ machines.
%
Let $\mathcal{N}(\mu_{\set{U}|\set{D}}^{\mbox{\tiny\emph{PITC}}},\Sigma_{\set{U}\set{U}|\set{D}}^{\mbox{\tiny\emph{PITC}}})$ be  the predictive Gaussian distribution computed by the centralized partially independent training conditional (PITC) approximation of FGP model \citep{candela05} where\vspace{-1mm}
\begin{equation} 
  \mu_{\set{U}|\set{D}}^{\mbox{\tiny\emph{PITC}}} \triangleq\mu_\set{U}+\Gamma_{\set{U}\set{D}}\left(\Gamma_{\set{D}\set{D}}+\Lambda\right)^{-1}\hspace{-0mm}\left(y_\set{D}-\mu_\set{D} \right)
  \label{eq:pitc.mu}\vspace{-1mm}
\end{equation}   
\begin{equation}   
\Sigma_{\set{U}\set{U}|\set{D}}^{\mbox{\tiny\emph{PITC}}} \triangleq\Sigma_{\set{U}\set{U}}-\Gamma_{\set{U}\set{D}}\left(\Gamma_{\set{D}\set{D}}+\Lambda\right)^{-1}\Gamma_{\set{D}\set{U}}
  \label{eq:pitc.var} \vspace{-0mm}
\end{equation} 
such that \vspace{-0mm}
\begin{equation} 
  \Gamma_{\set{B}\set{B}'} \triangleq\Sigma_{\set{B}\set{S}}\Sigma_{\set{S}\set{S}}^{-1}\Sigma_{\set{S}\set{B}'} 
  \label{eq:sparseq}
\end{equation} 
and 
$\Lambda$
is a 
block-diagonal matrix constructed from the $M$ diagonal blocks of $\Sigma_{\set{D}\set{D}|\set{S}}$,
each of which is a 
matrix $\Sigma_{\set{D}_m \set{D}_m|\set{S}}$
for $m=1,\ldots,M$ where $\set{D}=\bigcup_{m=1}^M \set{D}_m$.
Then, $\widehat{\mu}_{\set{U}}\hspace{-0mm}=\hspace{-0mm}\mu_{\set{U}|\set{D}}^{\mbox{\tiny\emph{PITC}}}$ and $\widehat{\Sigma}_{\set{U}\set{U}}\hspace{-0mm}=\hspace{-0mm}\Sigma_{\set{U}\set{U}|\set{D}}^{\mbox{\tiny\emph{PITC}}}$.
\label{thm:ppitc}
\end{thm}
\ifthenelse{\value{sol}=1}{The proof of Theorem~\ref{thm:ppitc} is previously reported in \citep{chen12} and reproduced in Appendix A to reflect our notations.\vspace{0mm}} 
{The proof of Theorem~\ref{thm:ppitc} is previously reported in \citep{chen12} and reproduced in Appendix A of \cite{AA13} to reflect our notations.\vspace{0mm}} 

\noindent \emph{Remark}. Since PITC generalizes the Bayesian Committee Machine (BCM) of \cite{Tresp02}, $p$PITC generalizes parallel BCM \citep{Ingram10}, the latter of which assumes the support set $\set{S}$ to be $\set{U}$ \citep{candela05}. As a result, parallel BCM does not scale well with large $\set{U}$.

Though $p$PITC scales very well with large data (Table~\ref{tab:compare}), it can predict poorly due to (a) 
loss of information caused by summarizing the realized outputs and correlation
structure of the original data; and (b) sparse coverage of $\set{U}$ by the support set.
We propose a novel parallel Gaussian process regression method called $p$PIC that combines the best of both worlds, that is, the predictive power of FGP and time efficiency of $p$PITC.
$p$PIC is based on the following intuition: 
A machine can exploit its local data to improve the predictions of the unobserved outputs  that are highly correlated with its data.
At the same time, $p$PIC can preserve the time efficiency of $p$PITC by exploiting its idea of encapsulating information into local and global summaries.
\begin{defi}[$p$PIC] Given a common support set
$\set{S}\subset\set{X}$
known to all $M$ machines, the global summary $(\ddot{y}_\set{S}, \ddot{\Sigma}_{\set{S}\set{S}})$, the local summary
$(\dot{y}_\set{S}^m, \dot{\Sigma}_{\set{S}\set{S}}^m)$, and the local data $(\set{D}_m, y_{\set{D}_m})$, each machine $m$ computes a predictive Gaussian distribution
$\mathcal{N}(\widehat{\mu}^+_{\set{U}_m},\widehat{\Sigma}^+_{\set{U}_m\set{U}_m})$ of the unobserved outputs for the set $\set{U}_m$ of inputs where 
\begin{equation}
%
%
\displaystyle \widehat{\mu}^+_{\set{U}_m}\triangleq \mu_{\set{U}_m} 
+\left({\Phi}_{\set{U}_m \set{S}}^{m}\ddot{\Sigma}_{\set{S}\set{S}}^{-1}\ddot{y}_\set{S}
-\Sigma_{\set{U}_m\set{S}}\Sigma_{\set{S}\set{S}}^{-1}\dot{y}_{\set{S}}^{m}\right)
+\dot{y}_{\set{U}_m}^{m}
%
\label{eq:lamu}
\end{equation}
\begin{equation}\vspace{-0mm}
%
\hspace{-1.8mm}
\begin{array}{c}
\widehat{\Sigma}^+_{\set{U}_m\set{U}_m}\hspace{-1.0mm}\triangleq\hspace{-0.5mm}  
\Sigma_{\set{U}_m\set{U}_m}
  -\left(\Phi_{\set{U}_m \set{S}}^{m}\Sigma_{\set{S}\set{S}}^{-1}\Sigma_{\set{S}\set{U}_m}
  -\Sigma_{\set{U}_m\set{S}}\Sigma_{\set{S}\set{S}}^{-1}\dot{\Sigma}_{\set{S}\set{U}_m}^{m}
  \right.
\\
  \hspace{11.3mm}\left.
  -\ {\Phi}_{\set{U}_m \set{S}}^{m}\ddot{\Sigma}_{\set{S}\set{S}}^{-1}{\Phi}_{\set{S}\set{U}_m}^{m}\right)
- \dot{\Sigma}_{\set{U}_m \set{U}_m}^{m}
%
\end{array}
\label{eq:lacov}
\end{equation}
such that
\begin{equation}
\Phi_{\set{U}_m \set{S}}^{m}\triangleq
\Sigma_{\set{U}_m \set{S}}+\Sigma_{\set{U}_m \set{S}}\Sigma_{\set{S}\set{S}}^{-1}\dot{\Sigma}_{\set{S}\set{S}}^{m}
-\dot{\Sigma}_{\set{U}_m \set{S}}^{m}
\label{eq:keys}
\end{equation}
and ${\Phi}_{\set{S}\set{U}_m}^{m}$ is the transpose of $\Phi_{\set{U}_m \set{S}}^{m}$.
\label{def:lap}
\end{defi}
\emph{Remark} $1$. The predictive Gaussian mean $\widehat{\mu}^+_{\set{U}_m}$  (\ref{eq:lamu}) and covariance $\widehat{\Sigma}^+_{\set{U}_m\set{U}_m}$ (\ref{eq:lacov}) of $p$PIC
exploit both summary information (i.e., bracketed term) and local information (i.e., last term). In contrast, $p$PITC only exploits the global summary (see (\ref{mupitc}) and (\ref{varpitc})).\vspace{0mm}  

\emph{Remark} $2$. To improve the predictive performance of $p$PIC, 
$\set{D}$ and $\set{U}$ should be partitioned
into tuples of $(\set{D}_1,\set{U}_1),\ldots,(\set{D}_M,\set{U}_M)$
such that the outputs $y_{\set{D}_m}$ and $Y_{\set{U}_m}$ are as highly correlated as possible for $m=1,\ldots,M$.
To achieve this, we employ a simple parallelized clustering scheme in our experiments: 
Each machine $m$ randomly selects a cluster center from its local data $\set{D_m}$ and informs the other machines about its chosen cluster center.
Then, each input in $\set{D}_m$ and $\set{U}_m$ is simply assigned to the ``nearest'' cluster center $i$ and sent to the corresponding machine $i$ while being subject to the constraints of the new $D_i$ and $U_i$ not exceeding $|\set{D}|/M$ and $|\set{U}|/M$, respectively.
More sophisticated clustering schemes can be utilized at the expense of greater time and communication complexity.\vspace{0mm}  

\emph{Remark} $3$. Predictive performances of $p$PITC and $p$PIC are improved by increasing size of $\set{S}$ at the expense of greater time, space, and communication complexity (Table~\ref{tab:compare}).
\begin{thm}
Let a common support set $\set{S}\subset\set{X}$ be known to all $M$ machines.
Let $\mathcal{N}(\mu_{\set{U}|\set{D}}^{\mbox{\tiny\emph{PIC}}},
    \Sigma_{\set{U}\set{U}|\set{D}}^{\mbox{\tiny\emph{PIC}}})$ be the predictive Gaussian
distribution computed by the centralized partially independent
conditional (PIC) approximation of FGP model \citep{snelson07} where 
\vspace{-0mm}
\begin{equation}
  \mu_{\set{U}|\set{D}}^{\mbox{\tiny\emph{PIC}}} \triangleq
\mu_\set{U}+\widetilde{\Gamma}_{\set{U}\set{D}}\left(
\Gamma_{\set{D}\set{D}}+\Lambda\right)^{-1}\hspace{-0mm}\left(y_\set{D}-\mu_\set{D} \right)\vspace{-3mm}
\label{eq:picmu} \end{equation}   
%
\vspace{-0mm}\begin{equation}\vspace{-0mm}
\Sigma_{\set{U}\set{U}|\set{D}}^{\mbox{\tiny\emph{PIC}}} \triangleq
\Sigma_{\set{U}\set{U}}-\widetilde{\Gamma}_{\set{U}\set{D}}\left(\Gamma_{\set{D}\set{D}}
+\Lambda\right)^{-1}\widetilde{\Gamma}_{\set{D}\set{U}}\vspace{-0mm}
\label{eq:picvar} \end{equation} 
and $\widetilde{\Gamma}_{\set{D}\set{U}}$ is the transpose of $\widetilde{\Gamma}_{\set{U}\set{D}}$ such that \vspace{-1mm}
%
%
%
\begin{equation}
\widetilde{\Gamma}_{\set{U}\set{D}}\triangleq \left(\widetilde{\Gamma}_{\set{U}_i\set{D}_{m}}\right)_{i, m = 1,\ldots,M} 
\label{eq:pick} \end{equation} 
\begin{equation}
\widetilde{\Gamma}_{\set{U}_i\set{D}_{m}}\triangleq\bigg{\{}
\begin{array}{cl}
  \Sigma_{\set{U}_i\set{D}_{m}}& \mbox{if}\ i=m,\\
  \Gamma_{\set{U}_i\set{D}_{m}}& \mbox{otherwise}.
\end{array} \label{eq:pickernel} \end{equation} 
%
%
Then, $\widehat{\mu}^+_\set{U} = \mu_{\set{U}|\set{D}}^{\mbox{\tiny\emph{PIC}}}$ and $\widehat{\Sigma}^+_{\set{U}\set{U}} = \Sigma_{\set{U}\set{U}|\set{D}}^{\mbox{\tiny\emph{PIC}}}$. 
\label{thm:ppic}
\end{thm}

\ifthenelse{\value{sol}=1}{The proof of Theorem~\ref{thm:ppic} is given in Appendix B.\vspace{0mm}} 
{Its proof is given in Appendix B of \cite{AA13}.\vspace{0mm}} 

\noindent
\emph{Remark} $1$. The equivalence results of Theorems~\ref{thm:ppitc}
and ~\ref{thm:ppic} imply that the computational load of the centralized PITC and PIC approximations of FGP can be distributed among $M$ parallel machines, hence improving the time efficiency and scalability of approximate GP regression (Table~\ref{tab:compare}).
\vspace{0mm}

\noindent \emph{Remark} $2$. The equivalence results also shed some light on the underlying properties of $p$PITC and $p$PIC based on the structural assumptions of PITC and PIC, respectively: 
$p$PITC assumes that $Y_{\set{D}_1},\ldots, Y_{\set{D}_M}, Y_{\set{U}_1},\ldots, Y_{\set{U}_M}$ are conditionally independent given $Y_\set{S}$.
In contrast, $p$PIC can predict the unobserved outputs $Y_{\set{U}}$ better since it imposes a less restrictive assumption of conditional independence between $Y_{\set{D}_1\bigcup \set{U}_1}, \ldots, Y_{\set{D}_M\bigcup \set{U}_M}$ given $Y_\set{S}$. 
This assumption further supports an earlier remark just before
    Theorem~\ref{thm:ppic} on clustering inputs $\set{D}_m$ and $\set{U}_m$ whose corresponding outputs are highly correlated for improving predictive performance of $p$PIC.
Experimental results on two real-world datasets (Section~\ref{sec:exp}) show that $p$PIC achieves predictive accuracy comparable to FGP and significantly better than $p$PITC, thus justifying the practicality of such an assumption. 
\section{Parallel Gaussian Process Regression using Incomplete Cholesky Factorization}
\label{sect:pgpi}
In this section, we will present another parallel Gaussian process called $p$ICF-based GP that  distributes the computational load among parallel machines to achieve efficient and scalable approximate GP regression by exploiting incomplete Cholesky factorization (ICF). A fundamental step of $p$ICF-based GP is to use ICF to approximate the covariance matrix $\Sigma_{\set{D}\set{D}}$ in (\ref{pmu}) and (\ref{pcov}) of FGP
by a low-rank symmetric positive semidefinite matrix: $\Sigma_{\set{D}\set{D}}\approx F^\top F + \sigma^2_n I$ where $F\in\mathbb{R}^{R\times|\set{D}|}$ denotes the upper triangular incomplete Cholesky factor and $R\ll |\set{D}|$ is the reduced rank.
The steps of performing $p$ICF-based GP are as follows:

{\scshape Step 1: Distribute data among $M$ machines.}

This step is the same as that of $p$PITC and $p$PIC in Section~\ref{sect:pgps}.

{\scshape Step 2: Run parallel ICF to produce incomplete Cholesky factor and distribute its storage.}

ICF can in fact be parallelized: Instead of using a column-based parallel ICF \citep{Golub96}, our proposed $p$ICF-based GP employs a row-based parallel ICF, the latter of which incurs lower time, space, and communication complexity. Interested readers are referred to \citep{Chang07} for a detailed implementation of the row-based parallel ICF, which is beyond the scope of this paper.
More importantly, it produces an upper triangular incomplete Cholesky factor $F\triangleq(F_1\cdots F_M)$ and each submatrix $F_m\in\mathbb{R}^{R\times|\set{D_m}|}$ is stored distributedly on machine $m$ for $m=1,\ldots,M$.

{\scshape Step 3: Each machine constructs and sends local summary to master.}

\begin{defi}[Local Summary] Given the local data $(\set{D}_m, y_{\set{D}_m})$ and incomplete Cholesky factor $F_m$, 
the local summary of machine $m$ is defined as a tuple $(\dot{y}_m, \dot{\Sigma}_m,\Phi_m)$ where
\begin{equation}
\dot{y}_{m}\triangleq F_m (y_{\set{D_m}}-\mu_{\set{D_m}})
\end{equation}
\begin{equation}
\dot{\Sigma}_{m}\triangleq F_m \Sigma_{\set{D_m}\set{U}}
\label{eq:lsvar}
\end{equation}
\begin{equation}
\Phi_m \triangleq F_m F^\top_m\ .\vspace{-2mm}
\end{equation}
 \label{def:lsi}
\end{defi}

{\scshape Step 4: Master constructs and sends global summary to $M$ machines.}

\begin{defi}[Global Summary] Given the local summary $(\dot{y}_m, \dot{\Sigma}_m,\Phi_m)$ of every machine $m = 1,\ldots,M$, the global summary is defined as a tuple $(\ddot{y},\ddot{\Sigma})$ where
\begin{equation}
\ddot{y}\triangleq \Phi^{-1}\sum^M_{m=1} \dot{y}_{m}
\label{eq:gsimu}
\end{equation}
\begin{equation}
\ddot{\Sigma}\triangleq\Phi^{-1}\sum^M_{m=1} \dot{\Sigma}_{m}
\label{eq:gsivar}
\end{equation}
such that $\Phi\triangleq I+\sigma^{-2}_n\sum^M_{m=1} \Phi_m$.
\label{def:gsi}
\end{defi}
\emph{Remark}. If $\sset{U}$ is large, the computation of (\ref{eq:gsivar}) can be parallelized by partitioning $\set{U}$: Let $\dot{\Sigma}_{m}\triangleq(\dot{\Sigma}^1_{m}\cdots\dot{\Sigma}^M_{m})$ where $\dot{\Sigma}^i_{m}\triangleq F_m \Sigma_{\set{D_m}\set{U}_i}$ is defined in a similar way as (\ref{eq:lsvar}) and $\sset{U}_i=|\set{U}|/M$. 
So, in Step $3$, instead of sending $\dot{\Sigma}_{m}$ to the master, each machine $m$ sends $\dot{\Sigma}^i_{m}$ to machine $i$ for $i=1,\ldots,M$.
Then, each machine $i$ computes and sends
$\ddot{\Sigma}_i \triangleq\Phi^{-1}\sum^M_{m=1} \dot{\Sigma}^i_{m}$ to every other machine to obtain $\ddot{\Sigma}=(\ddot{\Sigma}_1\cdots\ddot{\Sigma}_M)$.

{\scshape Step 5: Each machine constructs and sends predictive component to master.}

\begin{defi}[Predictive Component] Given the local data $(\set{D}_m, y_{\set{D}_m})$, a component $\dot{\Sigma}_m$ of the local summary, and the global summary $(\ddot{y},\ddot{\Sigma})$,
the predictive component of machine $m$ is defined as a tuple $(\widetilde{\mu}^m_{\set{U}}, \widetilde{\Sigma}^m_{\set{U}\set{U}})$ where
\begin{equation}
\widetilde{\mu}^m_{\set{U}}\triangleq \sigma^{-2}_n \Sigma_{\set{U}\set{D_m}}(y_{\set{D_m}}-\mu_{\set{D_m}}) - \sigma^{-4}_n \dot{\Sigma}^\top_m \ddot{y}
\label{eq:pcimu}
\end{equation}
\begin{equation}
\widetilde{\Sigma}^m_{\set{U}\set{U}}\triangleq \sigma^{-2}_n \Sigma_{\set{U}\set{D_m}}\Sigma_{\set{D_m}\set{U}} - \sigma^{-4}_n \dot{\Sigma}^\top_m \ddot{\Sigma}\ .
\label{eq:pcivar}
\end{equation}
\label{def:lpc}
\end{defi}

{\scshape Step 6: Master performs predictions.}

\begin{defi}[$p$ICF-based GP] Given the predictive component $(\widetilde{\mu}^m_{\set{U}}, \widetilde{\Sigma}^m_{\set{U}\set{U}})$ of every machine $m = 1,\ldots,M$, the master computes a predictive Gaussian distribution
$\mathcal{N}(\widetilde{\mu}_{\set{U}},\widetilde{\Sigma}_{\set{U}\set{U}})$ of the unobserved outputs for any set $\set{U}$ of inputs where 
\begin{equation}
\widetilde{\mu}_{\set{U}}\triangleq\mu_\set{U} + \sum^M_{m=1}\widetilde{\mu}^m_{\set{U}}
\label{eq:icfmu}
\end{equation}
\begin{equation}
\widetilde{\Sigma}_{\set{U}\set{U}}\triangleq\Sigma_\set{U\set{U}} - \sum^M_{m=1}\widetilde{\Sigma}^m_{\set{U}\set{U}}\ .
\label{eq:icfvar}
\end{equation}
\label{def:icfpred}
\end{defi}
\emph{Remark}. Predictive performance of $p$ICF-based GP can be improved by increasing rank $R$ at the expense of greater time, space, and communication complexity (Table~\ref{tab:compare}).
\begin{thm}
Let $\mathcal{N}(\mu^{\mbox{\tiny\emph{ICF}}}_{\set{U}|\set{D}},
\Sigma^{\mbox{\tiny\emph{ICF}}}_{\set{U}\set{U}|\set{D}})$ be the predictive Gaussian
distribution computed by the centralized ICF approximation of FGP model where 
\begin{equation}
\displaystyle\mu^{\mbox{\tiny\emph{ICF}}}_{\set{U}|\set{D}}\triangleq \mu_\set{U} + \Sigma_{\set{U}\set{D}}(F^\top F + \sigma^2_n I)^{-1}( y_\set{D} - \mu_\set{D} )
\label{imu}\vspace{-0mm}
\end{equation}
\begin{equation}
\displaystyle\Sigma^{\mbox{\tiny\emph{ICF}}}_{\set{U}\set{U}|\set{D}}\triangleq\Sigma_{\set{U}\set{U}}-\Sigma_{\set{U}\set{D}}(F^\top F + \sigma^2_n I)^{-1}\Sigma_{\set{D}\set{U}}\ .
\label{icov}\vspace{-0mm}
\end{equation}
Then, $\widetilde{\mu}_{\set{U}}=\mu^{\mbox{\tiny\emph{ICF}}}_{\set{U}|\set{D}}$ and $\widetilde{\Sigma}_{\set{U}\set{U}}=\Sigma^{\mbox{\tiny\emph{ICF}}}_{\set{U}\set{U}|\set{D}}$.
\label{thm:icfe}
\end{thm}
\ifthenelse{\value{sol}=1}{The proof of Theorem~\ref{thm:icfe} is given in Appendix C.} 
{Its proof is given in Appendix C of \cite{AA13}.} 

\noindent
\emph{Remark} $1$. The equivalence result of Theorem~\ref{thm:icfe} implies that the computational load of the centralized ICF approximation of FGP can be distributed among the $M$ parallel machines, hence improving the time efficiency and scalability of approximate GP regression (Table~\ref{tab:compare}).

\emph{Remark} $2$. By approximating the covariance matrix $\Sigma_{\set{D}\set{D}}$ in (\ref{pmu}) and (\ref{pcov}) of FGP with $F^\top F + \sigma^2_n I$, $\widetilde{\Sigma}_{\set{U}\set{U}}=\Sigma^{\mbox{\tiny\emph{ICF}}}_{\set{U}\set{U}|\set{D}}$ is not guaranteed to be positive semidefinite, hence rendering such a measure of predictive uncertainty not very useful.
However, it is observed in our experiments (Section~\ref{sec:exp}) that this problem can be alleviated by choosing a sufficiently large rank $R$.


\section{Analytical Comparison}
\label{sect:compare}
\begin{table*}
\caption{Comparison of time, space, and communication complexity between
  $p$PITC, $p$PIC, $p$ICF-based GP, PITC, PIC, ICF-based GP, and FGP. Note that PITC, PIC,
    and ICF-based GP are, respectively, the centralized counterparts of
      $p$PITC, $p$PIC, and $p$ICF-based GP, as proven in
      Theorems~\ref{thm:ppitc}, \ref{thm:ppic}, and~\ref{thm:icfe}.}
\centering
\begin{tabular}{l|ccc}
\hline
GP & Time complexity & \hspace{-1mm}Space complexity & \hspace{-0mm}Communication complexity\\
\hline
$p${PITC} & $\bigo{\displaystyle\sset{S}^2\left(\sset{S} + M + \frac{\sset{U}}{M}\right)+ \left(\frac{\sset{D}}{M}\right)^3}$ & \hspace{-1mm}$\bigo{\displaystyle\sset{S}^2+\left(\frac{\sset{D}}{M}\right)^2}$ & \hspace{-0mm}$\bigo{\sset{S}^2\log M}$ \\
$p${PIC} & $\bigo{\displaystyle \sset{S}^2\left(\sset{S} + M + \frac{\sset{U}}{M}\right) +  \left(\frac{\sset{D}}{M}\right)^3 + \sset{D}}$ & \hspace{-1mm}$\bigo{\displaystyle\sset{S}^2+\left(\frac{\sset{D}}{M}\right)^2}$ & \hspace{-0mm}$\bigo{\displaystyle\left(\sset{S}^2+\frac{\sset{D}}{M}\right)\log M}$ \\
$p${ICF-based} & $\bigo{\displaystyle R^2\left(R+M+\frac{\sset{D}}{M}\right)+R\sset{U}\left(M+\frac{\sset{D}}{M}\right)}$ & \hspace{-1mm}$\bigo{\displaystyle R^2+ R\frac{\sset{D}}{M}}$ & \hspace{-0mm}$\bigo{\displaystyle\left(R^2+R\sset{U}\right)\log M}$\\
PITC & $\bigo{\displaystyle\sset{S}^2\sset{D}+ \sset{D}\left(\frac{\sset{D}}{M}\right)^2}$ & \hspace{-1mm}$\bigo{\displaystyle\sset{S}^2+\left(\frac{\sset{D}}{M}\right)^2}$ & \hspace{-0mm}$-$\\
PIC & $\bigo{\displaystyle\sset{S}^2\sset{D}+ \sset{D}\left(\frac{\sset{D}}{M}\right)^2 + M\sset{D}}$ & \hspace{-1mm}$\bigo{\displaystyle\sset{S}^2+\left(\frac{\sset{D}}{M}\right)^2}$ & \hspace{-0mm}$-$\\
ICF-based & $\bigo{\displaystyle R^2\sset{D}+ R\sset{U}\sset{D}}$ & \hspace{-1mm}$\bigo{\displaystyle R\sset{D}}$ & \hspace{-0mm}$-$\\ 
FGP & $\bigo{\sset{D}^3}$ & \hspace{-1mm}$\bigo{\sset{D}^2}$ & \hspace{-0mm}$-$\\
\hline
\end{tabular}\vspace{-3mm}
\label{tab:compare}
\end{table*}
%
%
%
%
%
This section compares and contrasts the properties of the proposed parallel GPs analytically.
\subsection{Time, Space, and Communication Complexity} 
\label{sect:complexity}
Table~\ref{tab:compare} analytically compares the time, space, and communication complexity between $p$PITC, $p$PIC, $p$ICF-based GP, PITC, PIC, ICF-based GP, and FGP based on the following assumptions: (a) These respective methods compute the predictive means (i.e., $\widehat{\mu}_{\set{U}}$ (\ref{mupitc}), $\widehat{\mu}^+_{\set{U}}$ (\ref{eq:lamu}), $\widetilde{\mu}_{\set{U}}$ (\ref{eq:icfmu}), $ \mu_{\set{U}|\set{D}}^{\mbox{\tiny PITC}}$ (\ref{eq:pitc.mu}), $\mu_{\set{U}|\set{D}}^{\mbox{\tiny PIC}}$ (\ref{eq:picmu}), $\mu^{\mbox{\tiny ICF}}_{\set{U}|\set{D}}$ (\ref{imu}), and $\mu_{\set{U}|\set{D}}$ (\ref{pmu})) and their corresponding predictive variances (i.e., 
$\widehat{\Sigma}_{xx}$ (\ref{varpitc}), $\widehat{\Sigma}^+_{xx}$ (\ref{eq:lacov}), $\widetilde{\Sigma}_{xx}$ (\ref{eq:icfvar}), $\Sigma_{xx|\set{D}}^{\mbox{\tiny PITC}}$ (\ref{eq:pitc.var}), $\Sigma_{xx|\set{D}}^{\mbox{\tiny PIC}}$ (\ref{eq:picvar}), $\Sigma^{\mbox{\tiny ICF}}_{xx|\set{D}}$ (\ref{icov}), and $\Sigma_{xx|\set{D}}$ (\ref{pcov}) for all $x\in\set{U}$); (b) 
$\sset{U}<\sset{D}$ and recall $\sset{S},R\ll\sset{D}$; (c) the data is already distributed among $M$ parallel machines for $p$PITC, $p$PIC, and $p$ICF-based GP; and (d) for MPI, a broadcast operation in the communication network of $M$ machines incurs $\bigo{\log M}$ messages \citep{Grbovic07}. The observations are as follows:
%
\begin{enumerate}[(a)]
\item Our $p$PITC, $p$PIC, and $p$ICF-based GP improve the scalability of 
their centralized counterparts (respectively, PITC, PIC, and ICF-based GP) in the size $|\set{D}|$ of data by distributing their computational loads among the M parallel machines.
\item The speedups of $p$PITC, $p$PIC, and $p$ICF-based GP over their centralized counterparts deviate further from ideal speedup with increasing number $M$ of machines due to their additional $\set{O}(\sset{S}^2 M)$ or $\set{O}(R^2 M)$ time.
\item The speedups of $p$PITC and $p$PIC grow with increasing size $\sset{D}$ of data because, unlike the additional $\set{O}(\sset{S}^2 \sset{D})$ time of PITC and PIC that increase with more data, they do not have corresponding $\set{O}(\sset{S}^2 \sset{D}/M)$ terms. 
%
%
\item Our $p$PIC incurs additional $\bigo{\sset{D}}$ time and $\bigo{(\sset{D}/M)\log M}$-sized messages over $p$PITC due to its parallelized clustering (see Remark $2$ after Definition~\ref{def:lap}).
\item Keeping the other variables fixed, an increasing number $M$ of machines reduces the time and space complexity of $p$PITC and $p$PIC at a faster rate than $p$ICF-based GP while increasing size $\sset{D}$ of data raises the time and space complexity of $p$ICF-based GP at a slower rate than $p$PITC and $p$PIC.
\item Our $p$ICF-based GP distributes the memory requirement of ICF-based GP among the M parallel machines.
\item The communication complexity of $p$ICF-based GP depends on the number $\sset{U}$ of predictions whereas that of $p$PITC and $p$PIC are independent of it.
\end{enumerate}
\subsection{Online/Incremental Learning}
Supposing new data $(\set{D}',y_{\set{D}'})$ becomes available, $p$PITC and $p$PIC do not have to run Steps $1$ to $4$ (Section~\ref{sect:pgps}) on the entire data $(\set{D}\bigcup\set{D}',y_{\set{D}\bigcup\set{D}'})$.
The local and global summaries of the old data $(\set{D},y_{\set{D}})$ can in fact be reused and assimilated with that of the new data, thus saving the need of recomputing the computationally expensive matrix inverses in (\ref{eq:lsf}) and (\ref{eq:lsc}) for the old data. The exact mathematical details are omitted due to lack of space.
As a result, the time complexity of $p$PITC and $p$PIC can be greatly reduced in situations where new data is expected to stream in at regular intervals. In contrast, $p$ICF-based GP does not seem to share this advantage.
%
\subsection{Structural Assumptions}
The above advantage of online learning for $p$PITC and $p$PIC results
from their assumptions of conditional independence (see Remark $2$ after
    Theorem~\ref{thm:ppic}) given the support set.
With fewer machines, such an assumption is violated less, thus potentially improving their predictive performances.
In contrast, the predictive performance of $p$ICF-based GP is not affected by varying the number of machines. However, it suffers from a different problem: Utilizing a reduced-rank matrix approximation of $\Sigma_{\set{D}\set{D}}$, its resulting predictive covariance matrix $\widetilde{\Sigma}_{\set{U}\set{U}}$ is not guaranteed to be positive semidefinite (see Remark $2$ after Theorem~\ref{thm:icfe}), thus rendering such a measure of predictive uncertainty not very useful. 
It is observed in our experiments (Section~\ref{sec:exp}) that this problem can be alleviated by choosing a sufficiently large rank $R$.
\section{Experiments and Discussion}
\label{sec:exp}
This section empirically evaluates the predictive performances, time efficiency, scalability, and speedups of our proposed parallel GPs against their centralized counterparts and FGP
on two real-world datasets:
(a) The AIMPEAK dataset of size $\sset{D} = 41850$ contains traffic speeds (km/h) along $775$ road segments of an urban road network (including highways, arterials, slip roads, etc.) during the morning peak hours ($6$-$10$:$30$~a.m.) on April $20$, $2011$. 
The traffic speeds are the outputs.
The mean speed is $49.5$ km/h and the standard deviation is $21.7$ km/h.
Each input (i.e., road segment) is specified by a $5$-dimensional vector of features:
length, number of lanes, speed limit, direction, and time. The time
dimension comprises $54$ five-minute time slots. This spatiotemporal
traffic phenomenon is modeled using a relational GP (previously
    developed in \citep{chen12}) whose correlation structure can exploit
both the road segment features and road network topology information;
(b) The SARCOS dataset \citep{Vijayakumar05} of size $\sset{D} = 48933$ pertains to an inverse dynamics problem for a seven degrees-of-freedom SARCOS robot
arm. Each input denotes a $21$-dimensional vector of features: $7$ joint positions, $7$ joint velocities, and $7$ joint accelerations. Only one of the $7$ joint torques is used as the output. The mean torque is $13.7$ and the standard deviation is $20.5$. 

Both datasets are modeled using GPs whose prior covariance $\sigma_{xx'}$ is defined by
the squared exponential covariance function\footnote{For the AIMPEAK dataset, the domain of road segments is embedded into the Euclidean space using multi-dimensional scaling \citep{chen12} so that a squared exponential covariance function can then be applied.}:
\vspace{-1mm}
\begin{equation*}
\sigma_{xx'}\triangleq \sigma_s^2\exp\left({-\frac{1}{2}\sum_{i=1}^{d}
    \left(\frac{x_i-x'_i}{\ell_i}\right)^2}\right)+\sigma_n^2\delta_{xx'}
\label{eq:seard} \end{equation*}
where $x_i \left(x'_i \right)$ is the $i$-th component of the input
feature vector $x \left(x'\right)$, the hyperparameters
$\sigma^2_s,\sigma^2_n,\ell_1,\dots,\ell_d$ are, respectively, signal
variance, noise variance, and length-scales; and $\delta_{x x'}$ is a
Kronecker delta that is $1$ if $x = x'$ and $0$ otherwise.  The
hyperparameters are learned using randomly selected data of size $10000$
via maximum likelihood estimation \citep{rasmussen06}. 

For each dataset, 10\% of the data is randomly selected as test data for
predictions (i.e., as $\set{U}$). From the remaining data, training data
of varying sizes $\sset{D}= 8000$, $16000$, $24000$, and $32000$ are
randomly selected. The training data are distributed among $M$ machines
based on the simple parallelized clustering scheme in Remark $2$ after Definition~\ref{def:lap}.
Our $p$PITC and $p$PIC are evaluated using support sets of varying sizes
$\sset{S}=256$, $512$, $1024$, and $2048$ that are selected using differential entropy score criterion (see remark just after Definition~\ref{def:ls}).
Our $p$ICF-based GP is evaluated using varying reduced ranks $R$ of the same values as $\sset{S}$ in the AIMPEAK domain and twice the values of $\sset{S}$ in the SARCOS domain.
%

Our experimental platform is a cluster of $20$ computing nodes connected
via gigabit links: Each node runs a Linux system with
Intel$\circledR$ Xeon$\circledR$ CPU E$5520$ at $2.27$~GHz and $20$~GB memory.
Our parallel GPs are tested with different number $M= 4$, $8$, $12$, $16$, and $20$ of computing nodes.
\subsection{Performance Metrics}
The tested GP regression methods are evaluated with four
different performance metrics: 
(a) Root mean square error (RMSE) $\sqrt{\sset{U}^{-1}\sum_{x\in
\set{U}}\left(y_x-\mu_{x|\set{D}}\right)^2}$; (b) mean negative log
probability (MNLP) $0.5\sset{U}^{-1} \sum_{x\in \set{U}} \left(
(y_x-\mu_{x|\set{D}})^2/\Sigma_{xx|\set{D}} + \log
(2\pi\Sigma_{xx|\set{D}}) \right)$ \citep{rasmussen06}; (c) incurred
time; and (d) speedup is defined as the incurred time of a sequential/centralized algorithm divided by that of its corresponding parallel algorithm. For the first two
metrics, the tested methods have to plug their predictive mean and
variance into $\mu_{u|\set{D}}$ and $\Sigma_{uu|\set{D}}$, respectively. 
\subsection{Results and Analysis}
%
In this section, we analyze the results that are obtained by averaging over $5$ random instances.

\subsubsection{Varying size $\sset{D}$ of data} 
\begin{figure}[h!] \begin{tabular}{cc}
\vspace{-1mm} \hspace{-5mm}
\includegraphics[scale=0.22]{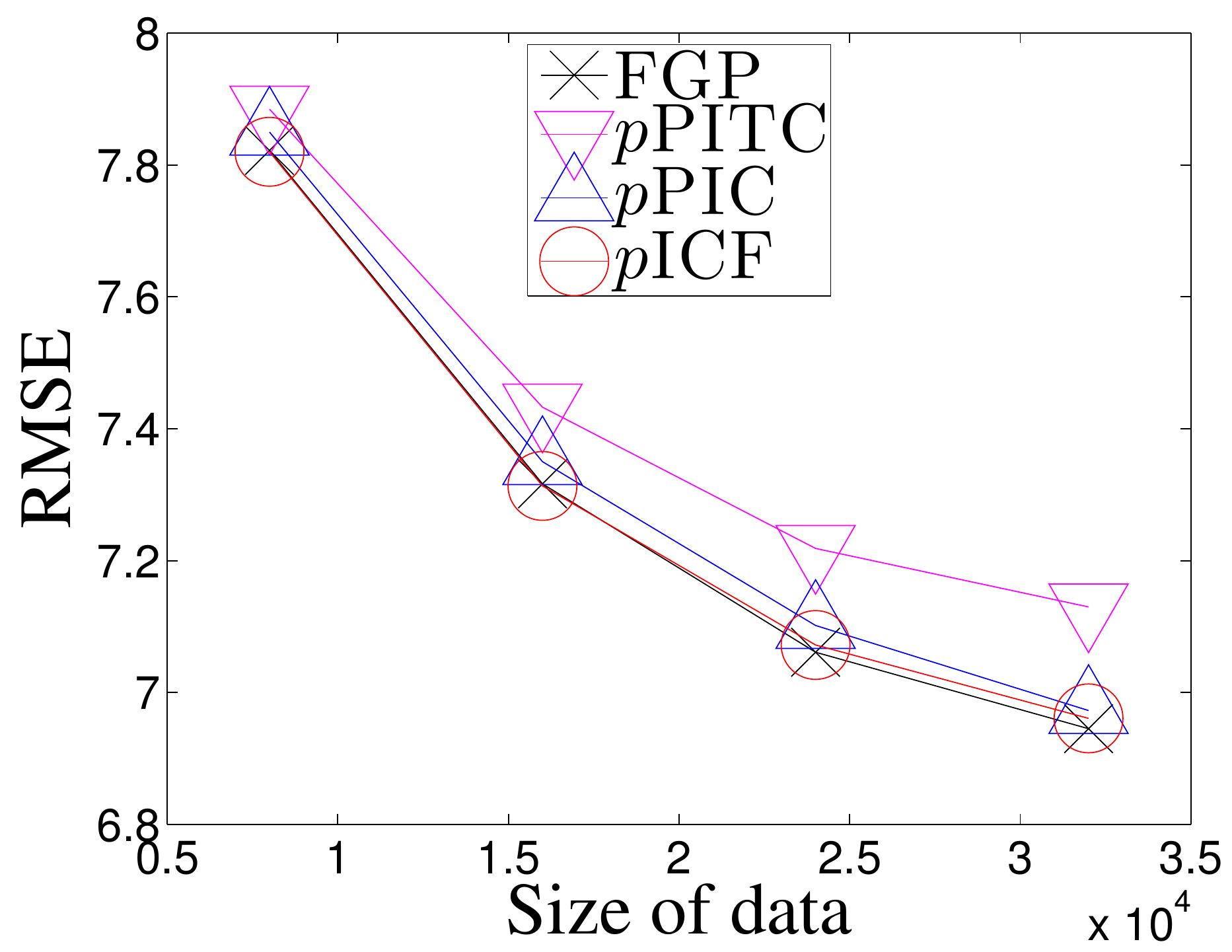}
& \hspace{-5mm}
\includegraphics[scale=0.22]{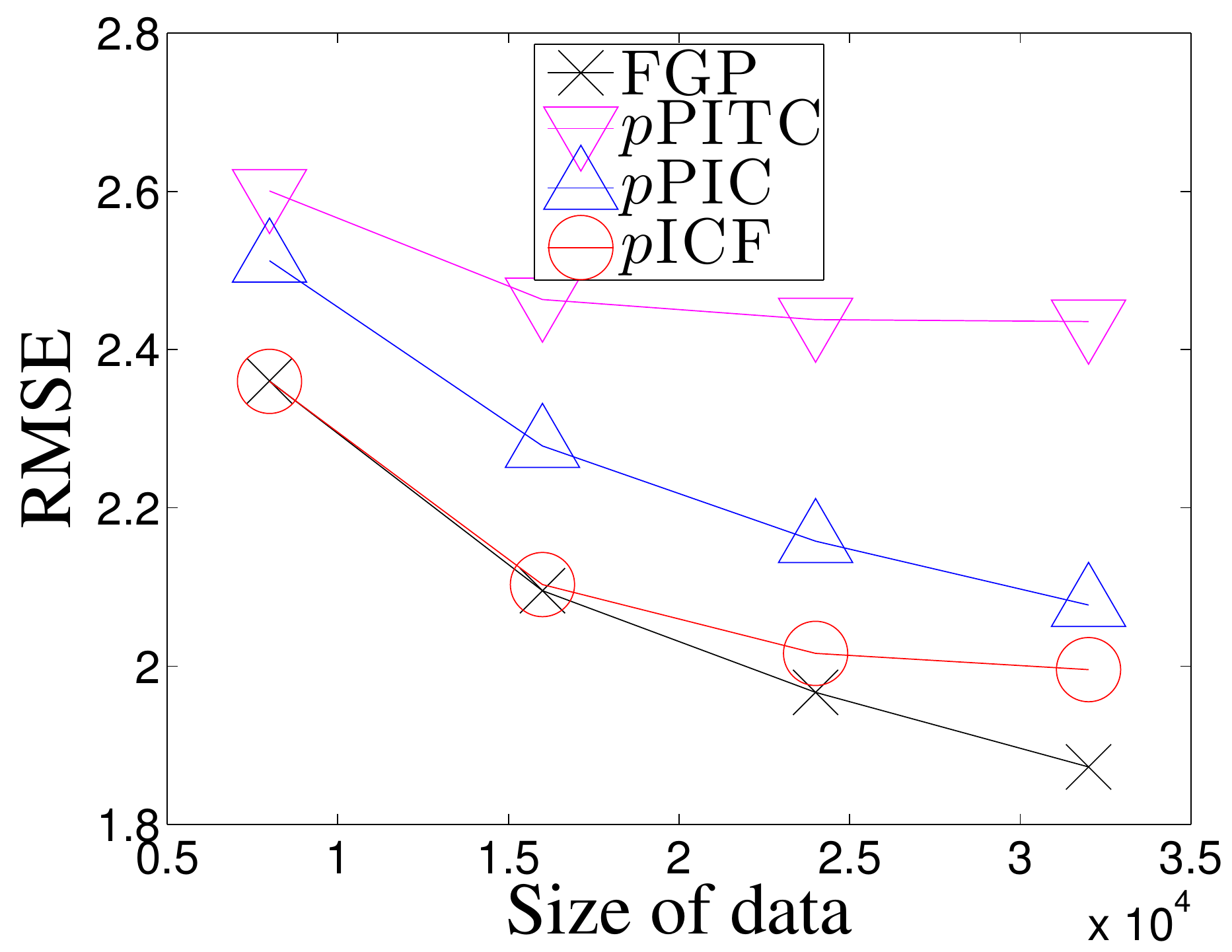}
\\
\vspace{-0mm} \hspace{-5mm} (a) AIMPEAK & \hspace{-5mm} (e) SARCOS\\ 
\vspace{-1mm} \hspace{-5mm}
\includegraphics[scale=0.22]{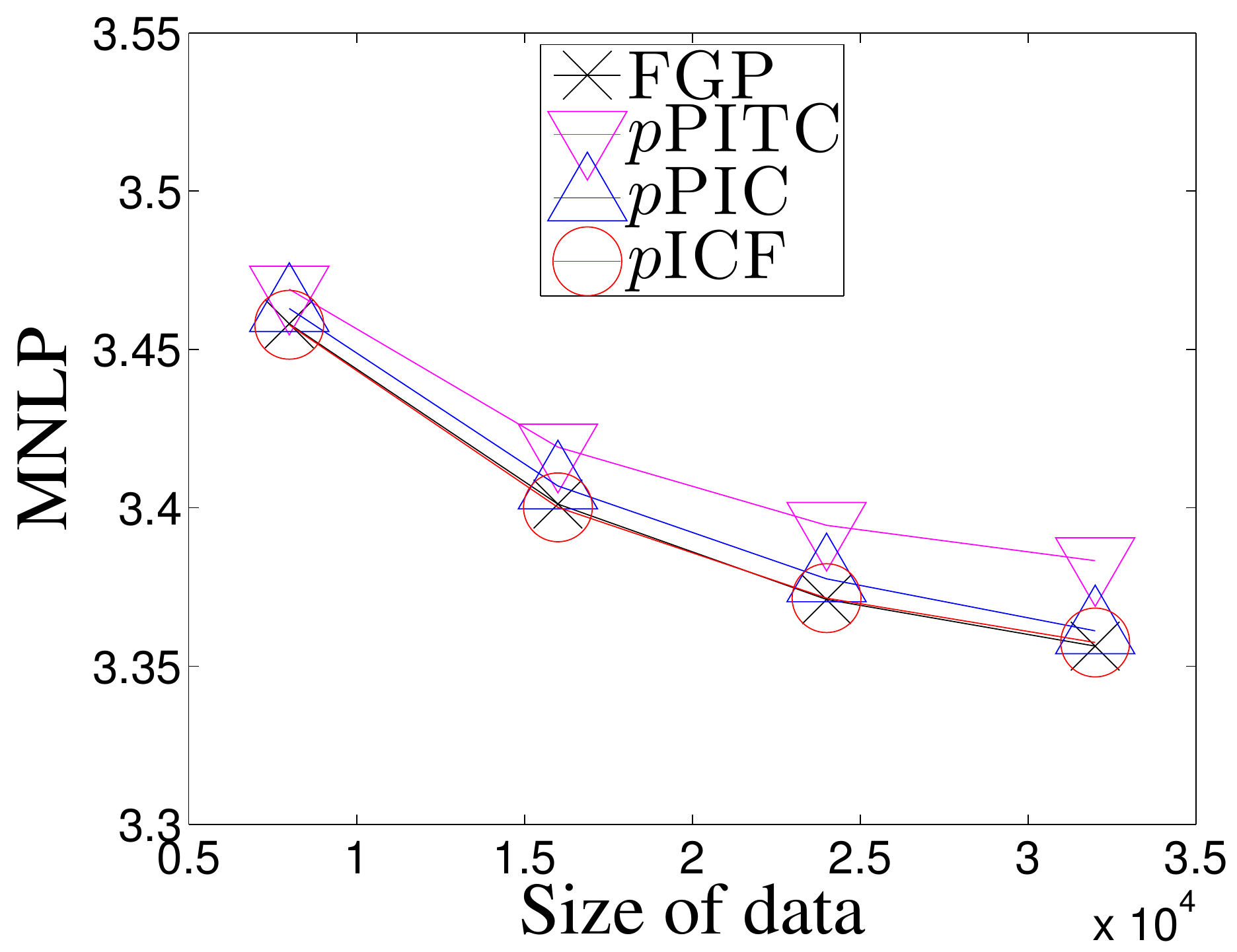}
& \hspace{-5mm}
\includegraphics[scale=0.22]{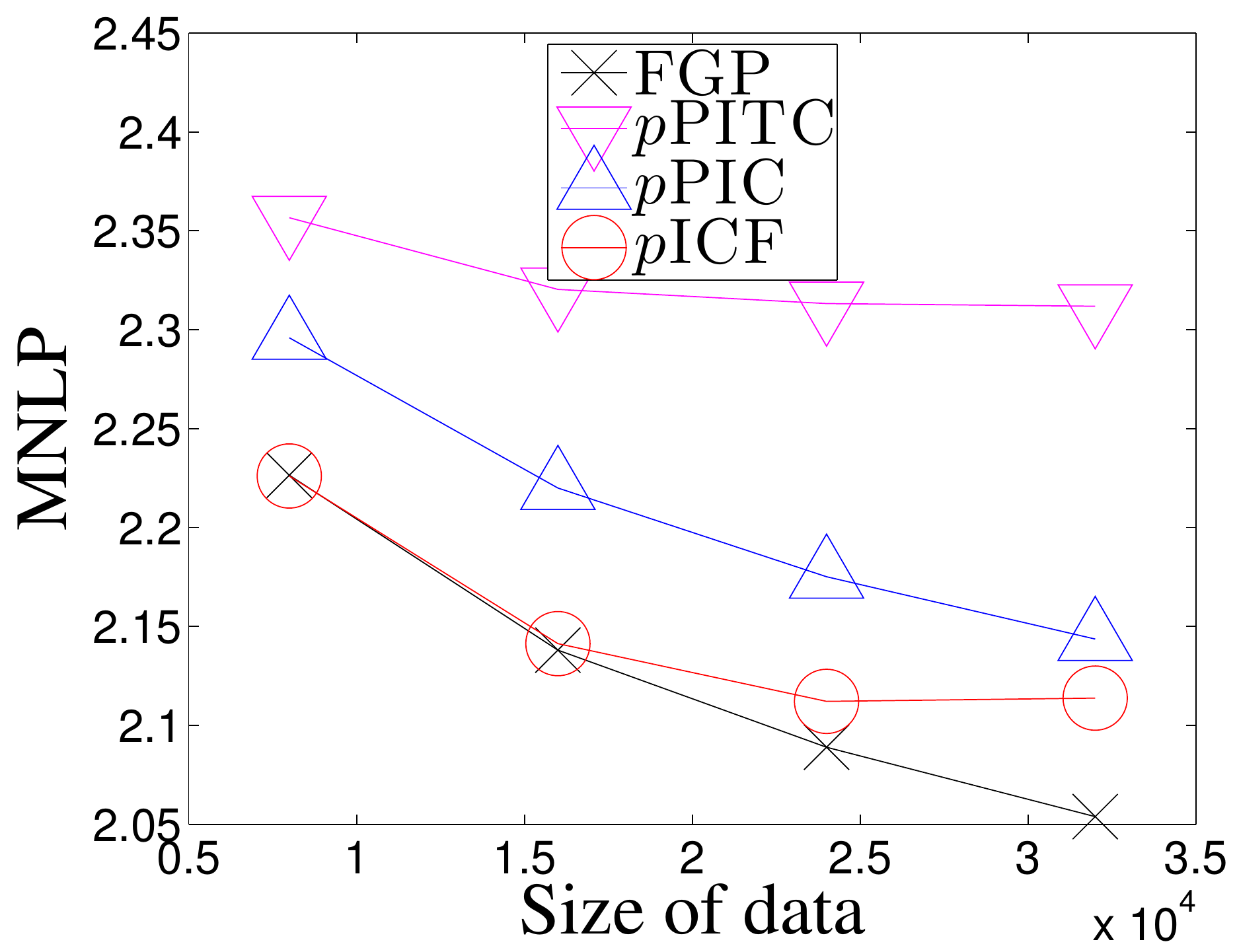}
\\
\vspace{-0mm} \hspace{-5mm} (b) AIMPEAK & \hspace{-5mm} (f) SARCOS\\ 
\vspace{-1mm} \hspace{-5mm}
\includegraphics[scale=0.22]{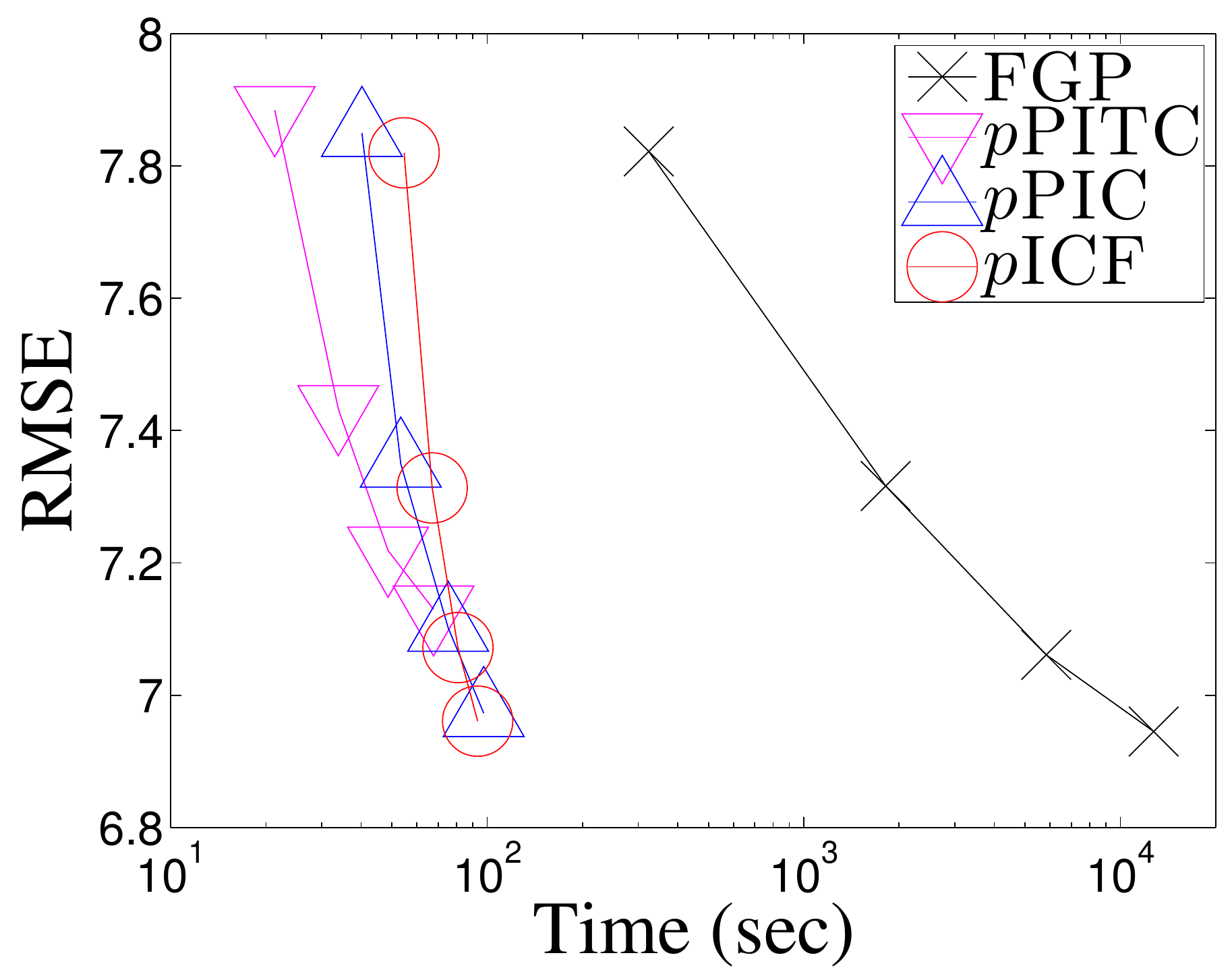}
& \hspace{-5mm}
\includegraphics[scale=0.22]{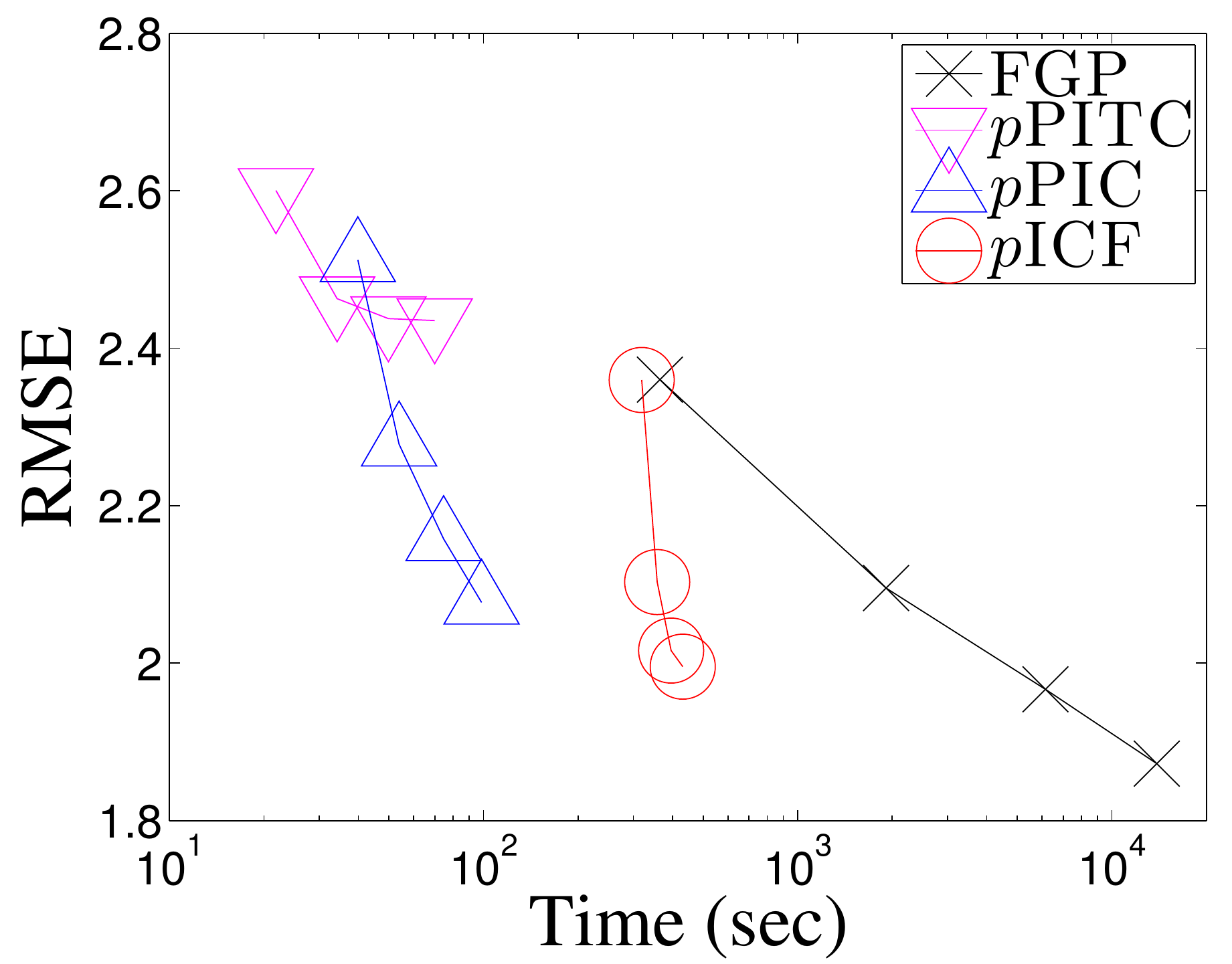}
\\
\vspace{-0mm} \hspace{-5mm} (c) AIMPEAK & \hspace{-5mm} (g) SARCOS\\ 
\vspace{-1mm} \hspace{-5mm}
\includegraphics[scale=0.22]{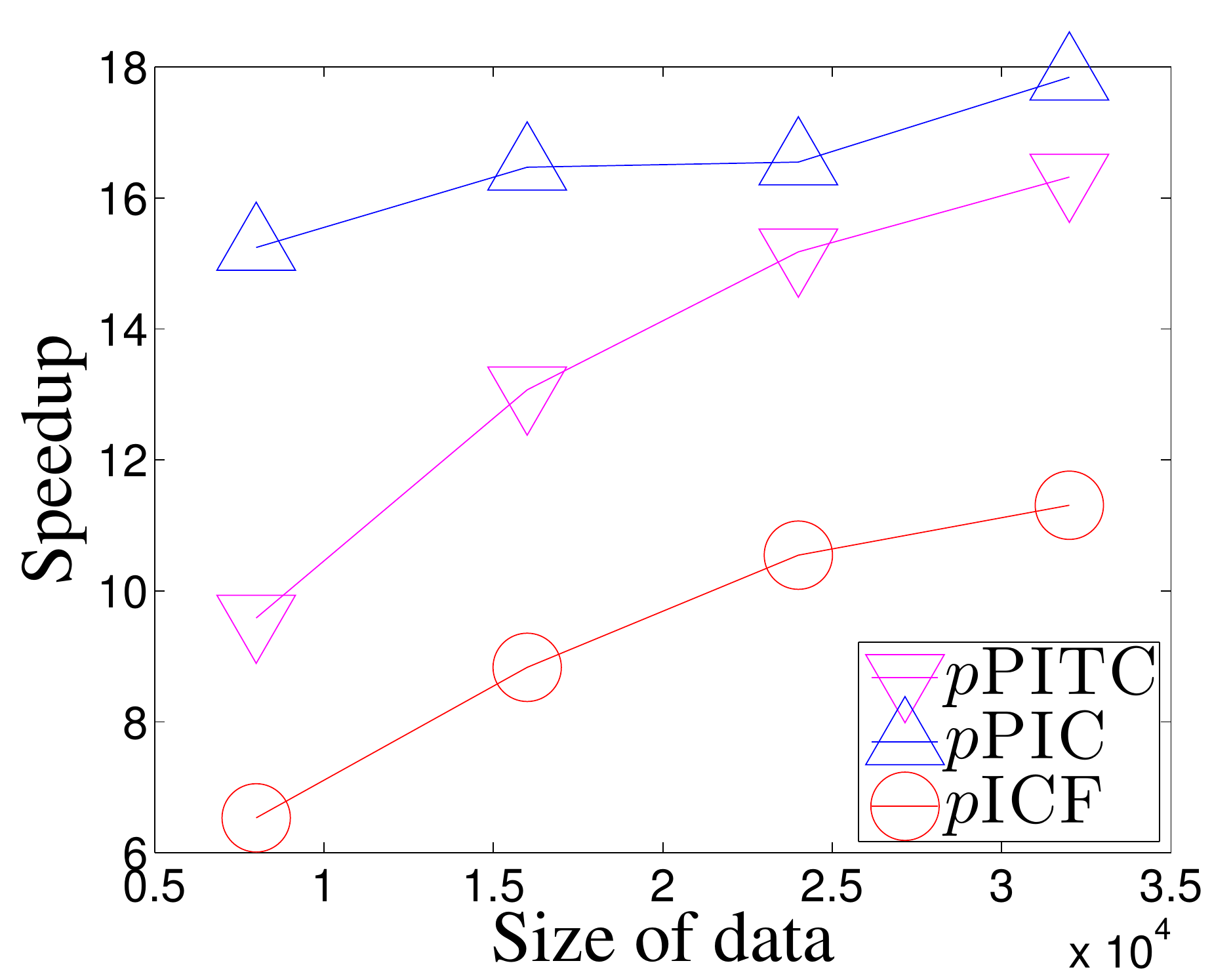}
& \hspace{-5mm}
\includegraphics[scale=0.22]{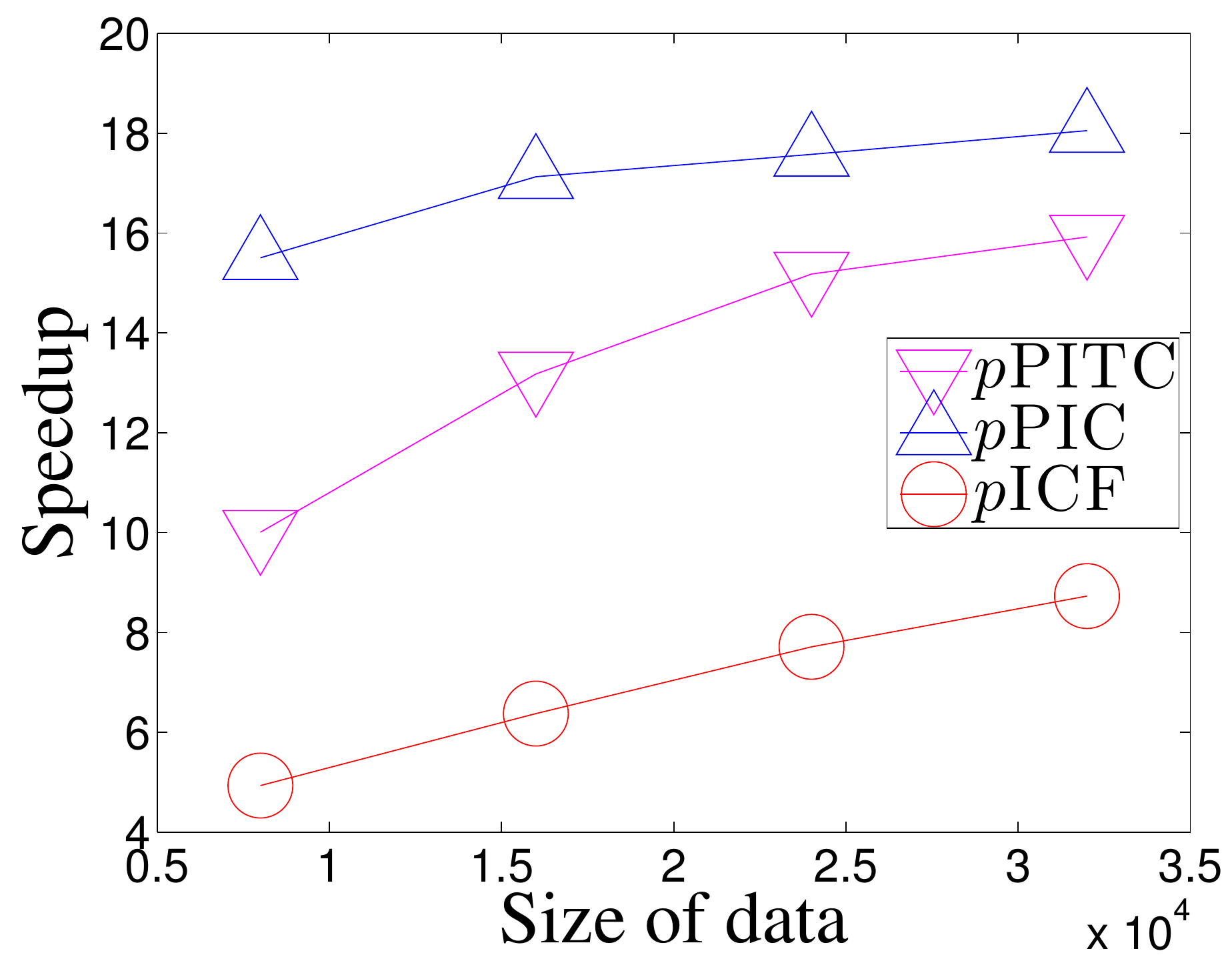}
\\
\vspace{-0mm} \hspace{-5mm} (d) AIMPEAK & \hspace{-5mm} (h) SARCOS\\ 
\vspace{-3mm}
%
\end{tabular}
\caption{Performance of parallel GPs with varying data sizes $\sset{D}=8000$, $16000$, $24000$, and $32000$, number $M=20$ of machines,
               support set size $\sset{S}=2048$, and reduced rank
                 $R=2048$ ($4096$) in the AIMPEAK (SARCOS) domain.\vspace{-0mm}}
\label{fig:varyd} \end{figure}

Figs.~\ref{fig:varyd}a-b and~\ref{fig:varyd}e-f show that the predictive performances of our parallel GPs improve with more data and are comparable to that of FGP, hence justifying the practicality of their inherent structural assumptions.
%

From Figs.~\ref{fig:varyd}e-f, it can be observed that the predictive performance of $p$ICF-based GP is very close to that of FGP when $\sset{D}$ is relatively small (i.e., $\sset{D}=8000$, $16000$). But, its performance approaches that of $p$PIC as $\sset{D}$ increases further because the reduced rank $R=4096$ of $p$ICF-based GP is not large enough (relative to $\sset{D}$) to maintain its close performance to FGP.
In addition, $p$PIC achieves better predictive performance than $p$PITC since the former can exploit local information (see Remark $1$ after Definition~\ref{def:lap}).

Figs.~\ref{fig:varyd}c and~\ref{fig:varyd}g indicate that our parallel GPs are significantly more time-efficient and scalable than FGP (i.e., $1$-$2$ orders of magnitude faster) while achieving comparable predictive performance.
Among the three parallel GPs, $p$PITC and $p$PIC are more time-efficient and thus more capable of meeting the real-time prediction requirement of a time-critical application/system.

Figs.~\ref{fig:varyd}d and~\ref{fig:varyd}h show that the speedups of our parallel
GPs over their centralized counterparts increase with more data, which agree with observation c in Section~\ref{sect:complexity}.
$p$PITC and $p$PIC achieve better speedups than $p$ICF-based GP. 
%
\begin{figure}[h!] \begin{tabular}{cc}
\vspace{-1mm} \hspace{-5mm}
\includegraphics[scale=0.22]{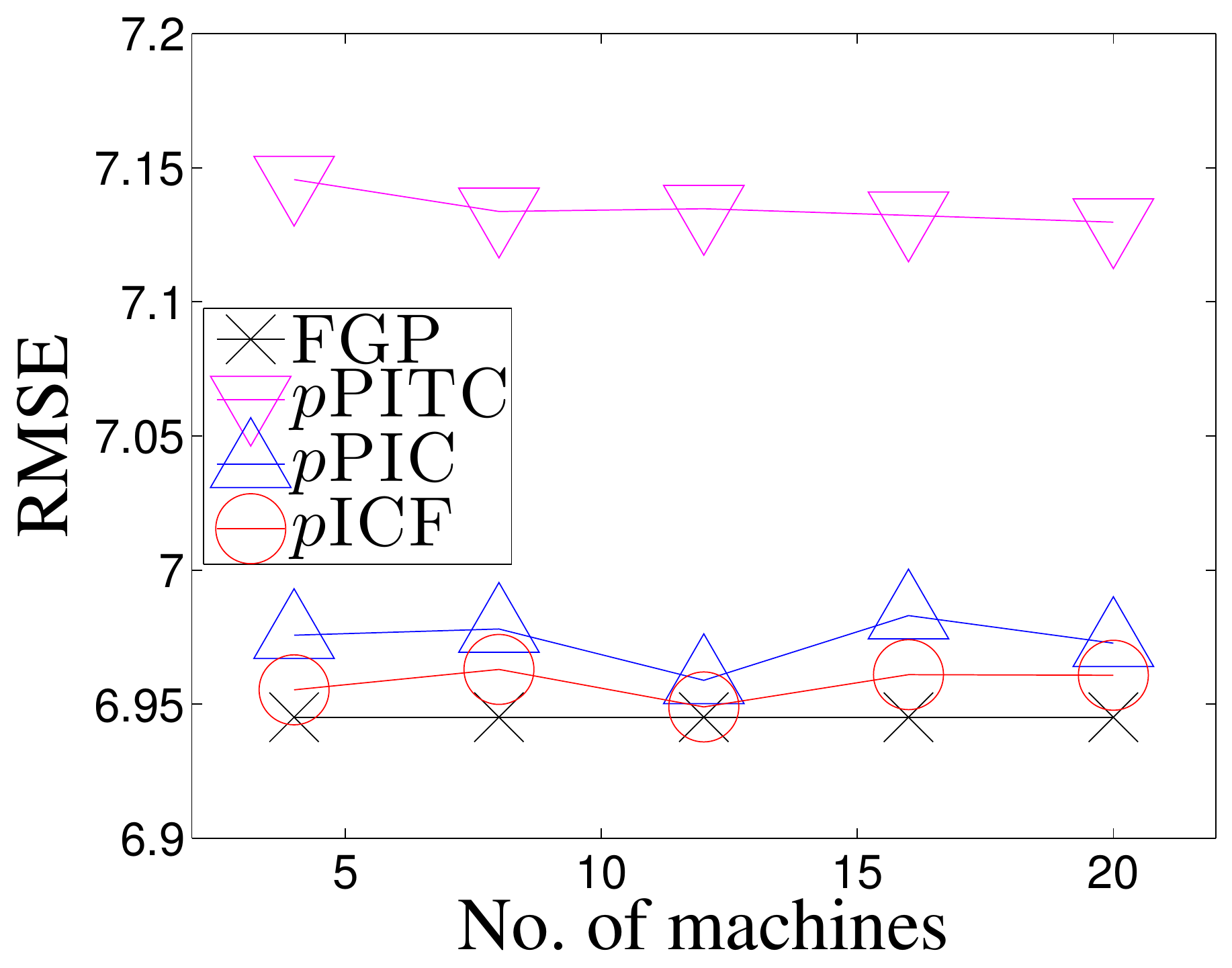}
& \hspace{-5mm}
\includegraphics[scale=0.22]{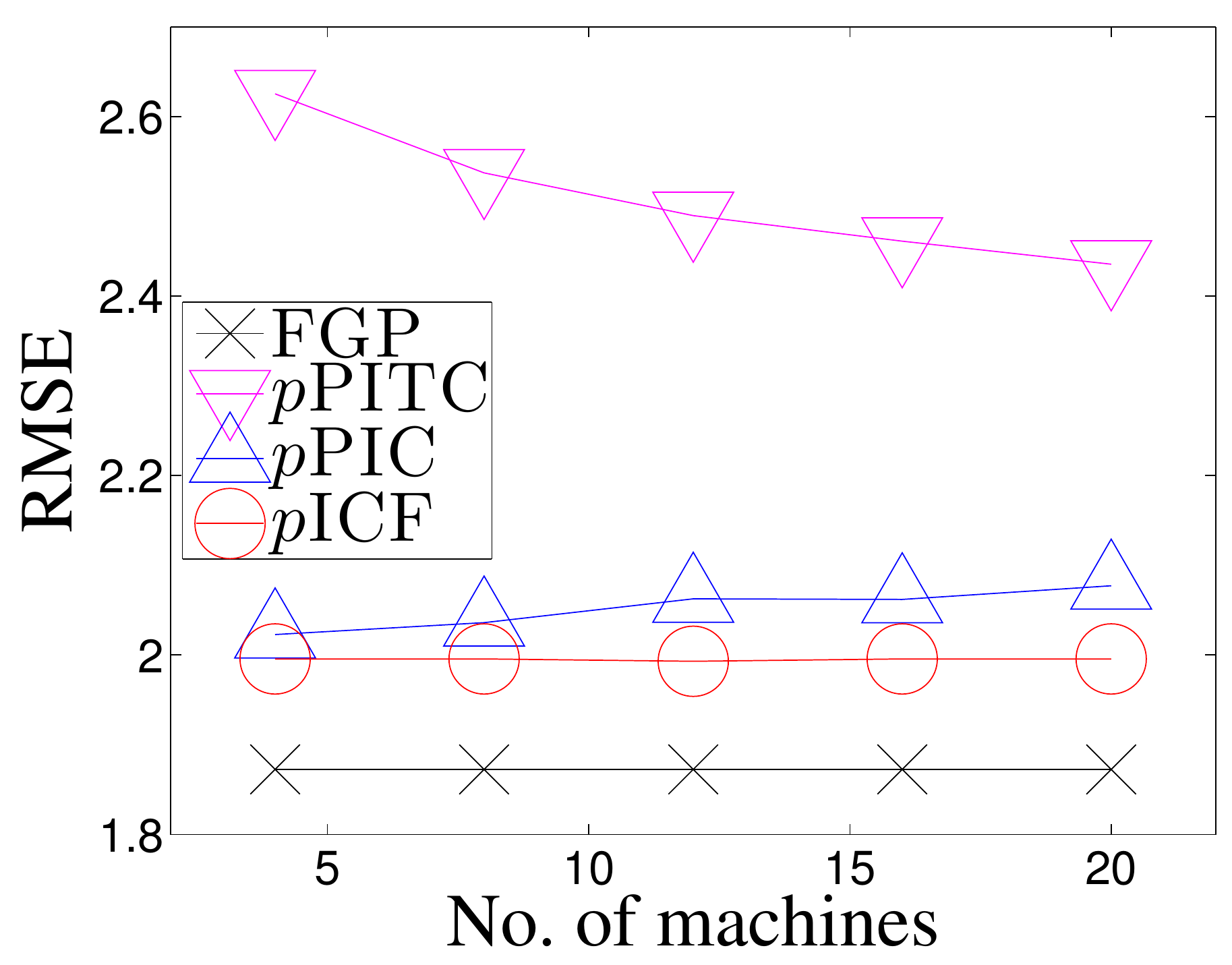}
\\
\vspace{-0mm} \hspace{-5mm} (a) AIMPEAK& \hspace{-5mm} (e) SARCOS\\ 
\vspace{-1mm} \hspace{-5mm}
\includegraphics[scale=0.22]{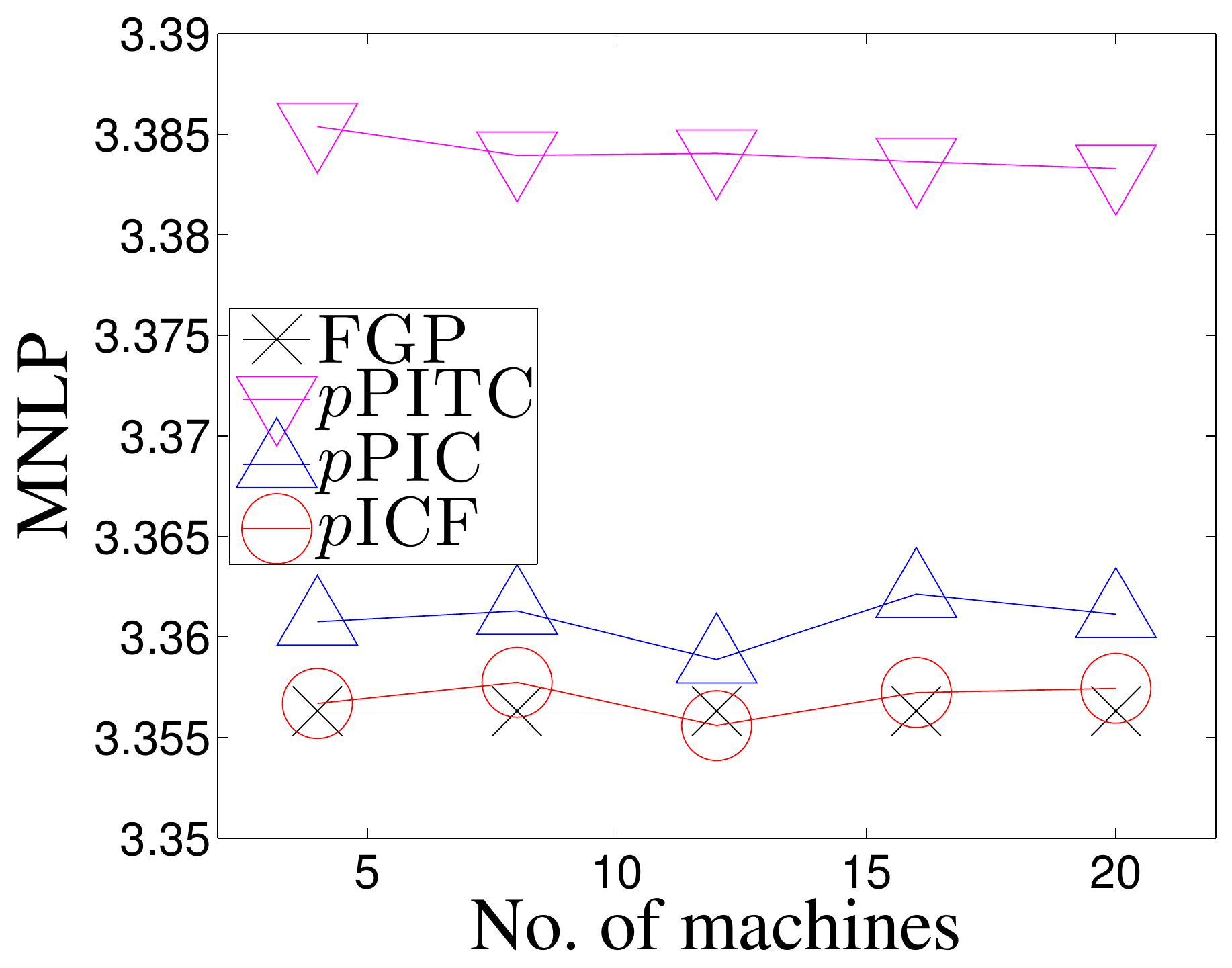}
& \hspace{-5mm}
\includegraphics[scale=0.22]{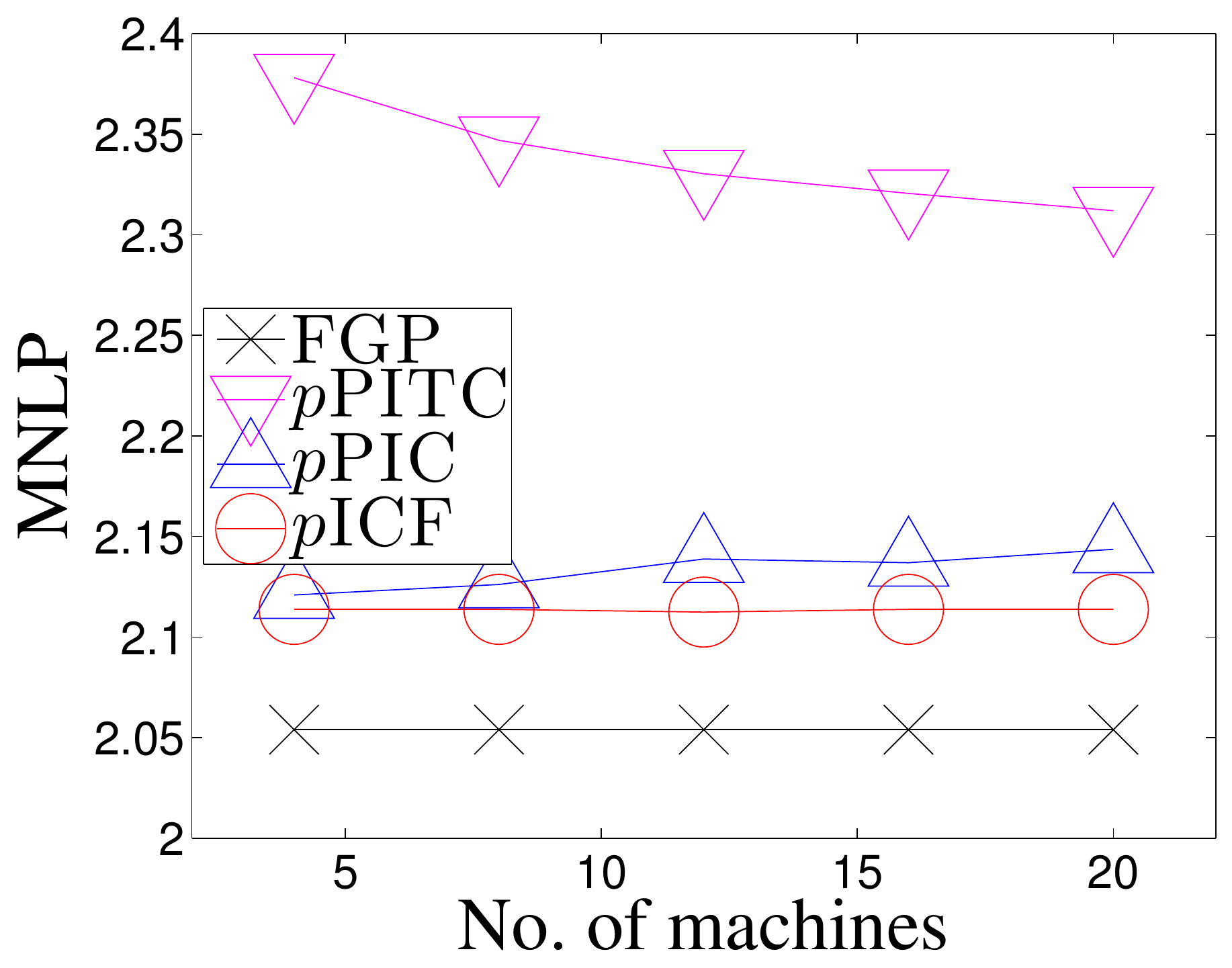}
\\
\vspace{-0mm} \hspace{-5mm} (b) AIMPEAK& \hspace{-5mm} (f) SARCOS\\ 
\vspace{-1mm} \hspace{-5mm}
\includegraphics[scale=0.22]{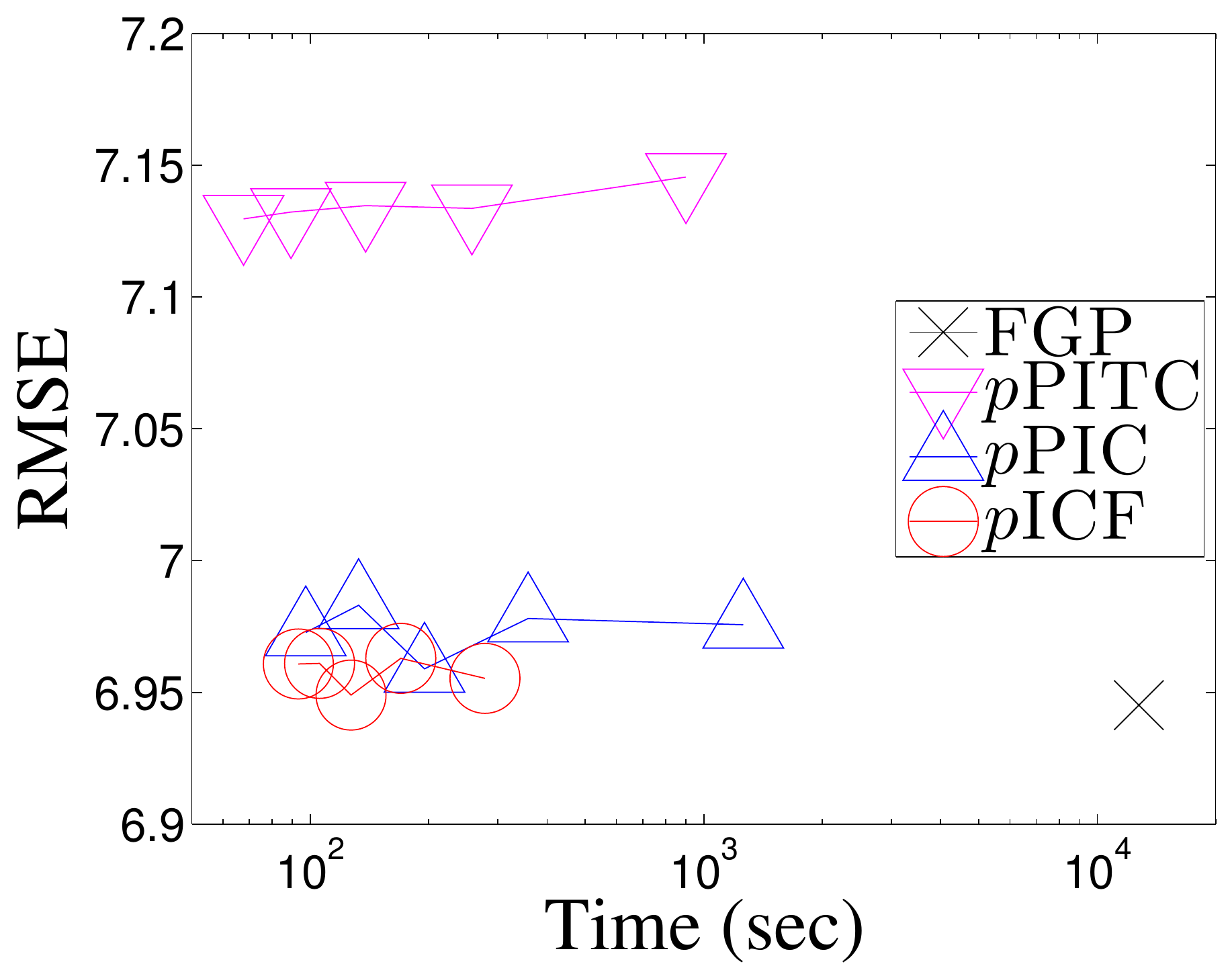}
& \hspace{-5mm}
\includegraphics[scale=0.22]{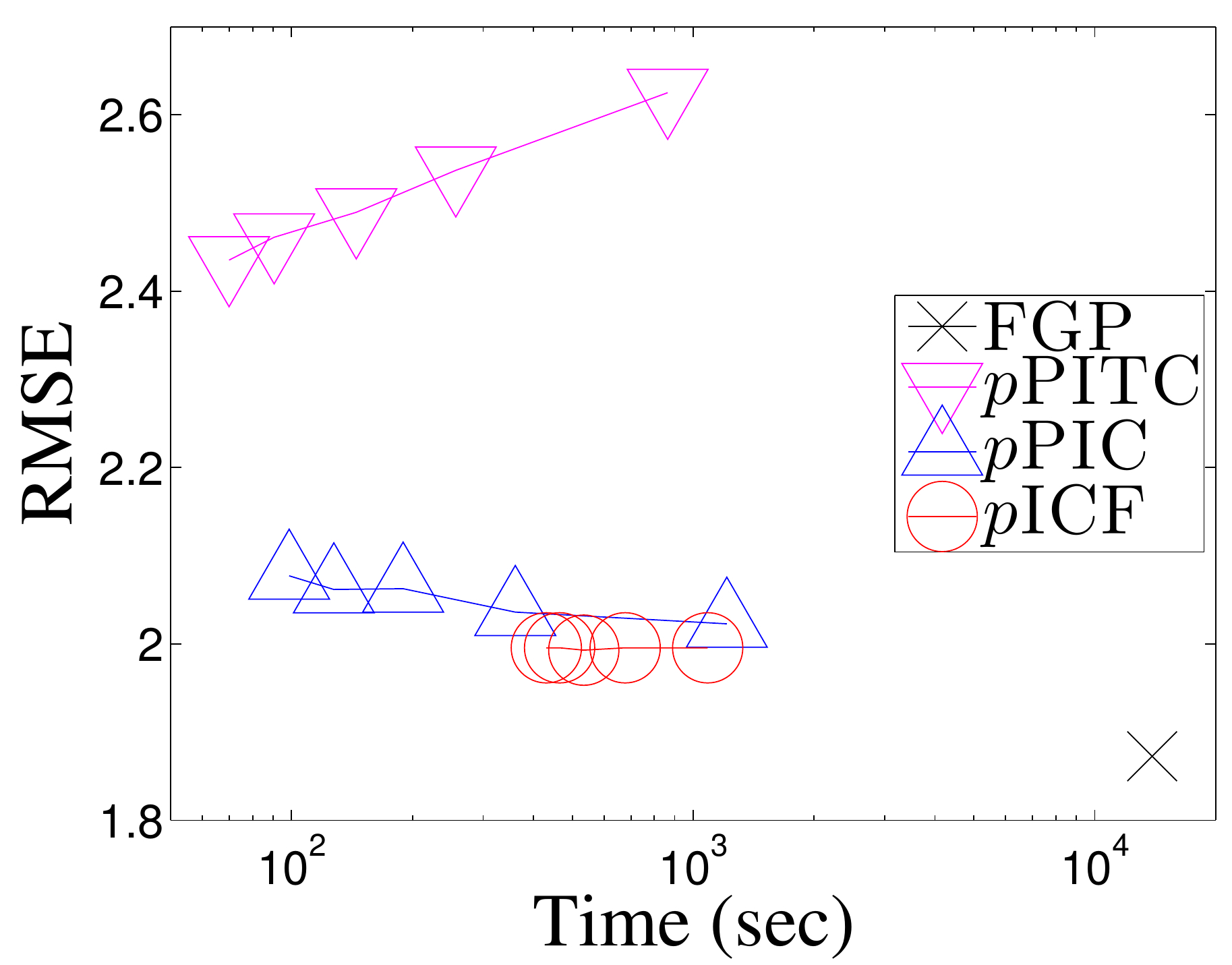}
\\
\vspace{-0mm} \hspace{-5mm} (c) AIMPEAK& \hspace{-5mm} (g) SARCOS\\ 
\vspace{-1mm} \hspace{-3mm}
\includegraphics[scale=0.22]{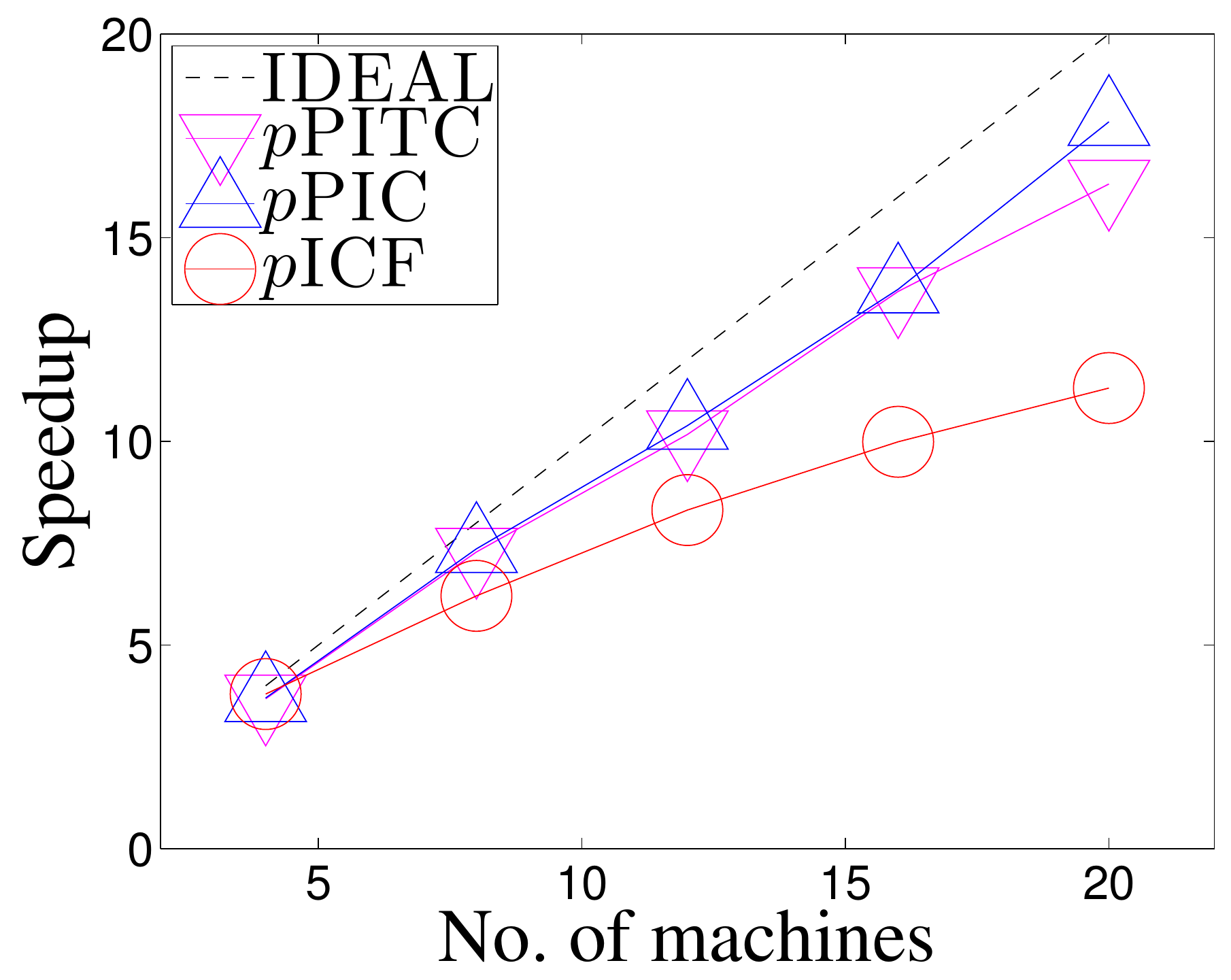}
& \hspace{-5mm}
\includegraphics[scale=0.22]{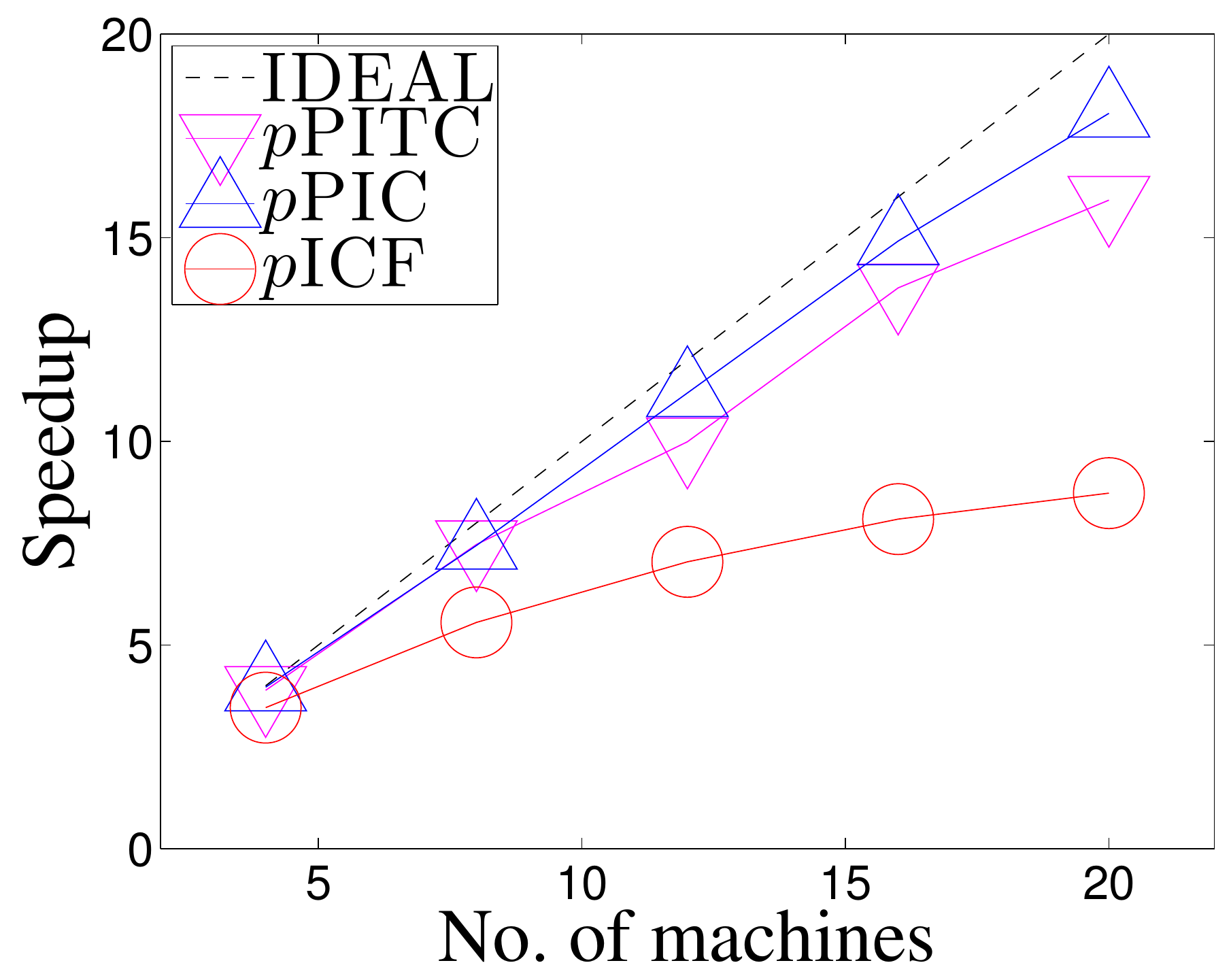}
\\
\vspace{-0mm} \hspace{-5mm} (d) AIMPEAK& \hspace{-5mm} (h) SARCOS\\ 
\vspace{-2mm}
%
\end{tabular}
\caption{Performance of parallel GPs with varying number $M=4$, $8$,
  $12$, $16$, $20$ of machines, data size $\sset{D}=32000$, support set
    size $\set{S}=2048$, and reduced rank $R=2048$ ($4096$) in the
    AIMPEAK (SARCOS) domain.  The \emph{ideal} speedup of a parallel algorithm
    is defined to be the number $M$ of machines running it.
    \vspace{-0mm}}

\label{fig:varyk} \end{figure}
\subsubsection{Varying number $M$ of machines}
Figs.~\ref{fig:varyk}a-b and~\ref{fig:varyk}e-f show that $p$PIC and $p$ICF-based GP
achieve predictive performance comparable to that of FGP with different
number $M$ of machines. $p$PIC achieves better predictive
performance than $p$PITC due to its use of local information (see Remark $1$ after Definition~\ref{def:lap}).  

From Figs.~\ref{fig:varyk}e-f, it can be observed that as the number $M$ of machines increases, the predictive performance of $p$PIC drops slightly due to smaller
size of local data $\set{D}_m$ assigned to each machine. 
In contrast, the predictive performance of $p$PITC improves: If the number $M$ of machines is small as compared to the actual number of clusters in the data, then the clustering scheme (see Remark $2$ after Definition~\ref{def:lap}) may assign data from different clusters to the same machine or data from the same cluster to multiple machines.
Consequently, the conditional independence assumption is violated.
Such an issue is mitigated by increasing the number $M$ of machines to
achieve better clustering, hence resulting in better predictive performance.

Figs.~\ref{fig:varyk}c and~\ref{fig:varyk}g show that $p$PIC and $p$ICF-based GP are significantly more time-efficient than FGP (i.e., $1$-$2$ orders of magnitude faster) while achieving comparable predictive performance. This is previously explained in the analysis of their time complexity (Table~\ref{tab:compare}).

Figs.~\ref{fig:varyk}c and~\ref{fig:varyk}g also reveal that as the number $M$ of machines increases, the incurred time of $p$PITC and $p$PIC decreases at a faster rate than that of $p$ICF-based GP, which agree with observation e in Section~\ref{sect:complexity}.
Hence, we expect $p$PITC and $p$PIC to be more time-efficient than $p$ICF-based GP when the number $M$ of machines increases beyond $20$.

Figs.~\ref{fig:varyk}d and~\ref{fig:varyk}h show that the speedups of our parallel
GPs over their centralized counterparts deviate further from the ideal speedup with a greater number $M$ of machines, which agree with observation b in Section~\ref{sect:complexity}. The speedups of $p$PITC and $p$PIC are closer  to the ideal speedup than that of $p$ICF-based GP. 
\subsubsection{Varying support set size $\sset{S}$ and reduced rank $R$}
\begin{figure}[h!] \begin{tabular}{cc}
\vspace{-1mm} \hspace{-5mm}
\includegraphics[scale=0.22]{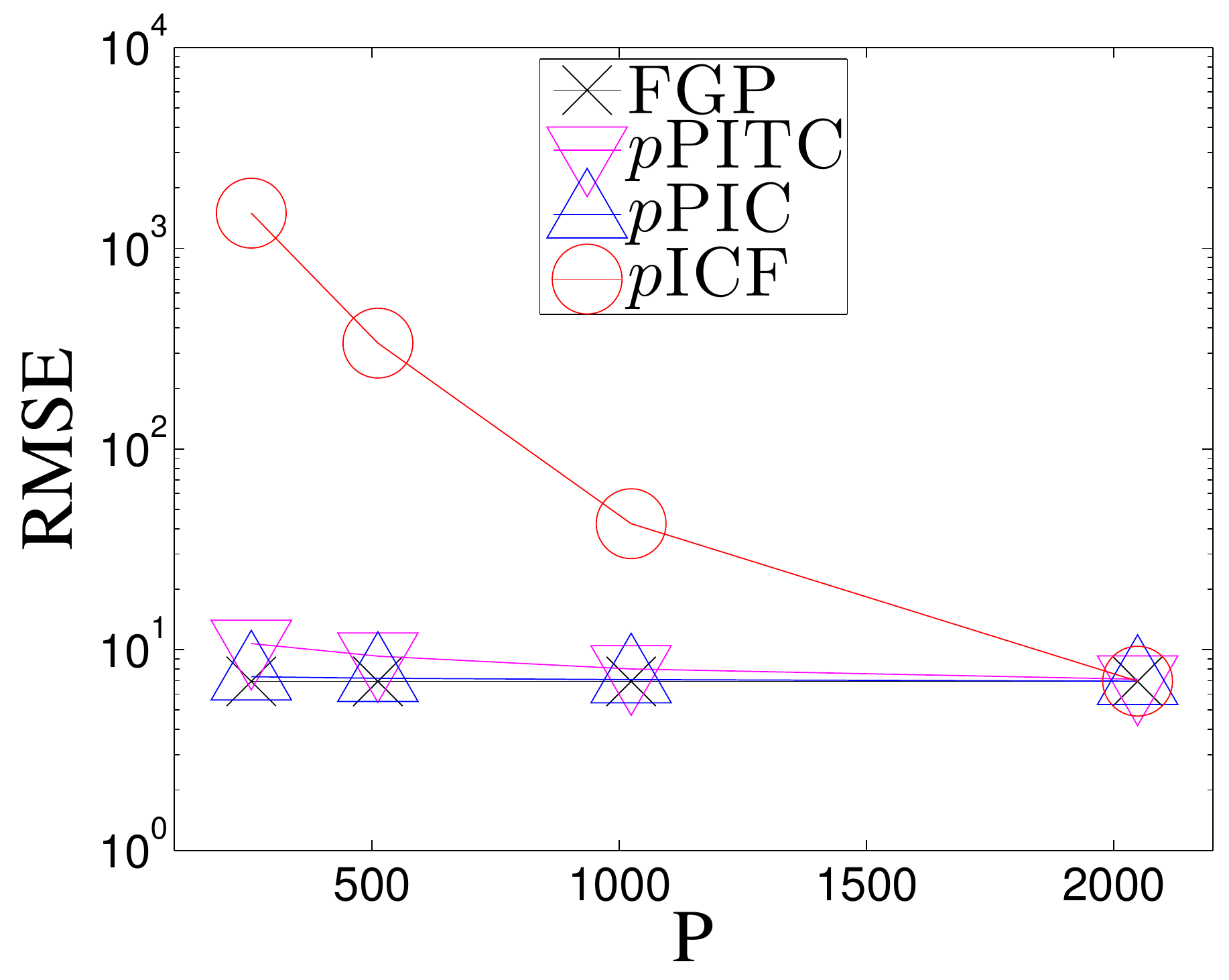} &
\hspace{-5mm}
\includegraphics[scale=0.22]{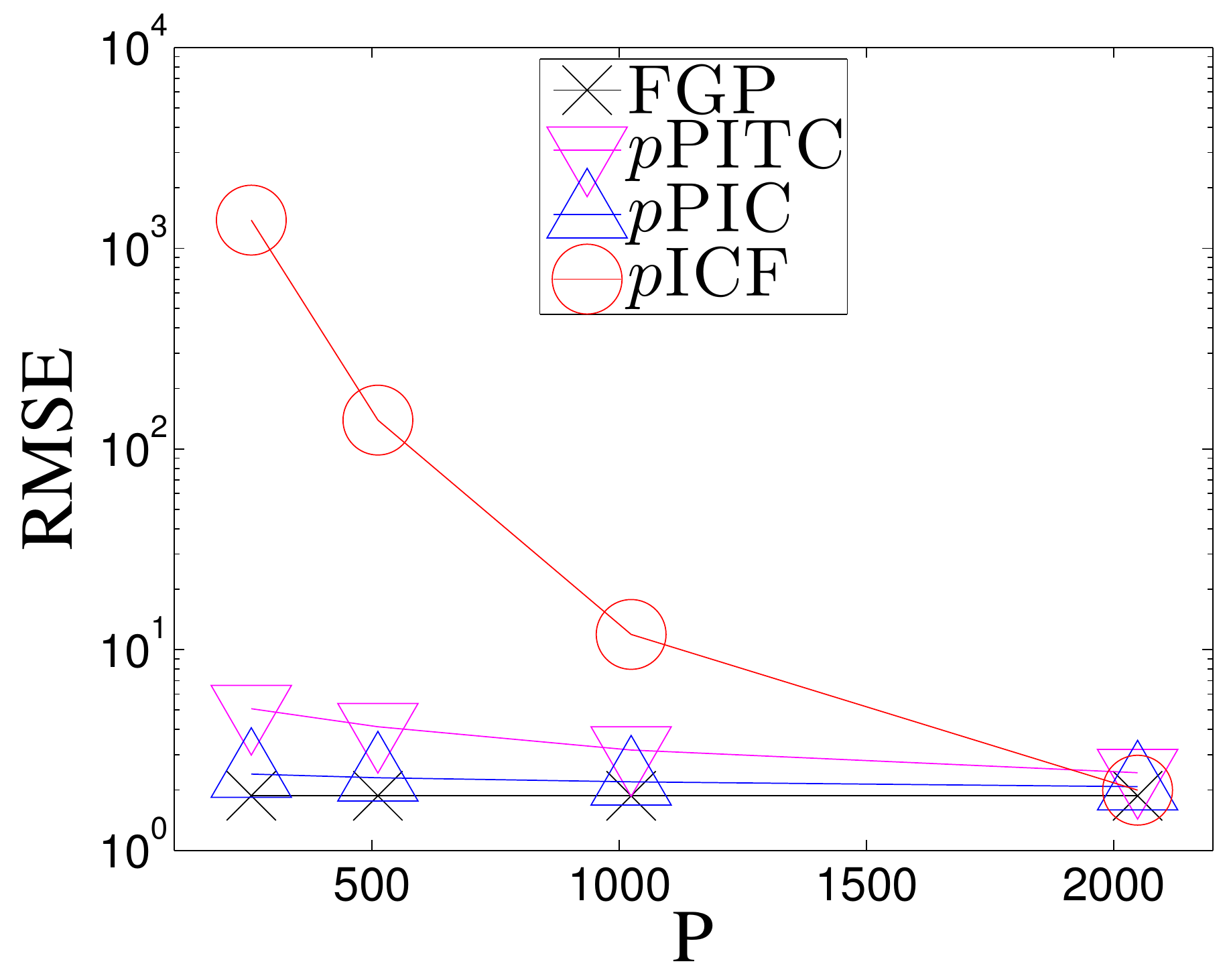} \\
\vspace{-0mm} \hspace{-5mm} (a) AIMPEAK& \hspace{-5mm} (e) SARCOS\\ 
\vspace{-1mm} \hspace{-5mm}
\includegraphics[scale=0.22]{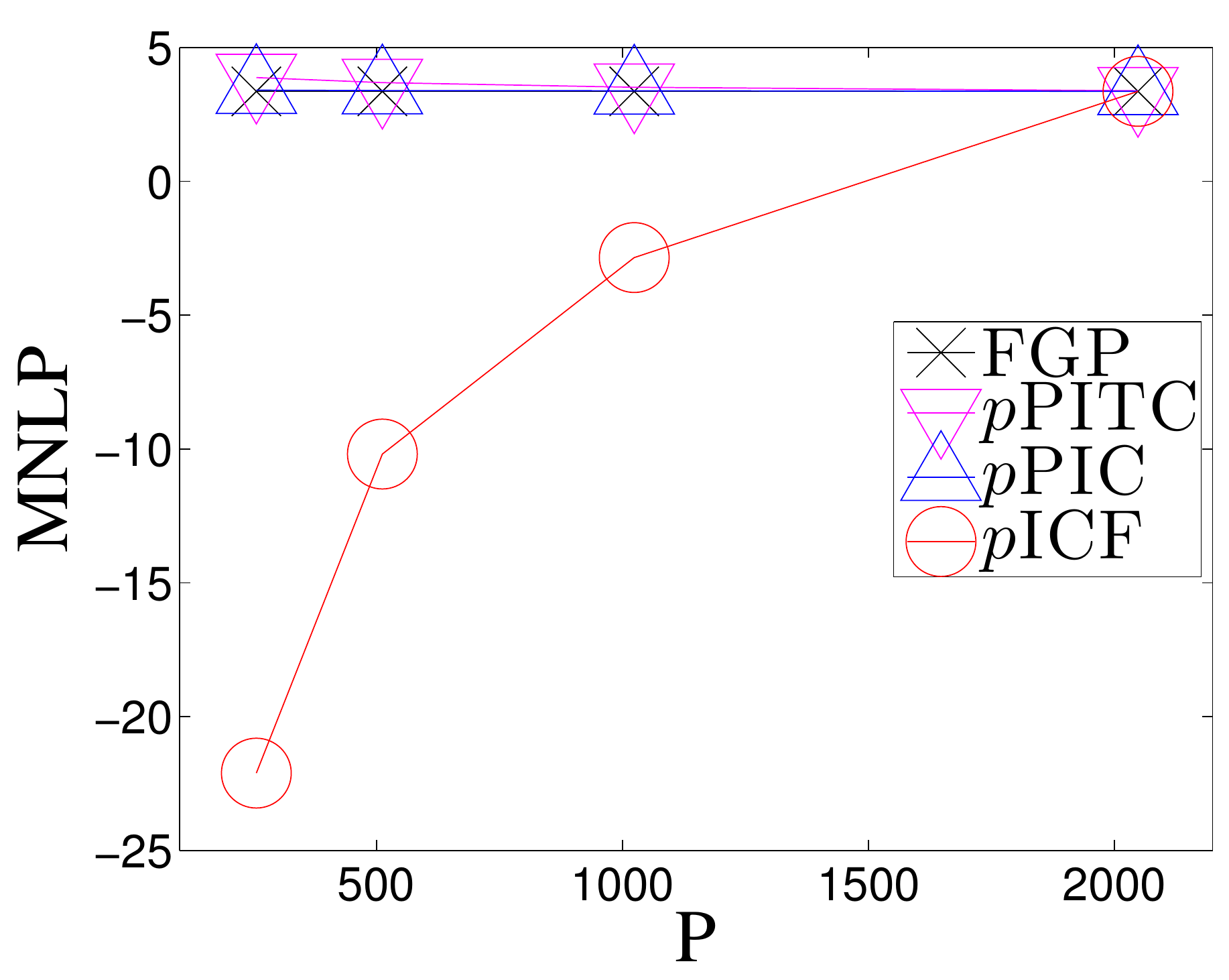} &
\hspace{-5mm}
\includegraphics[scale=0.22]{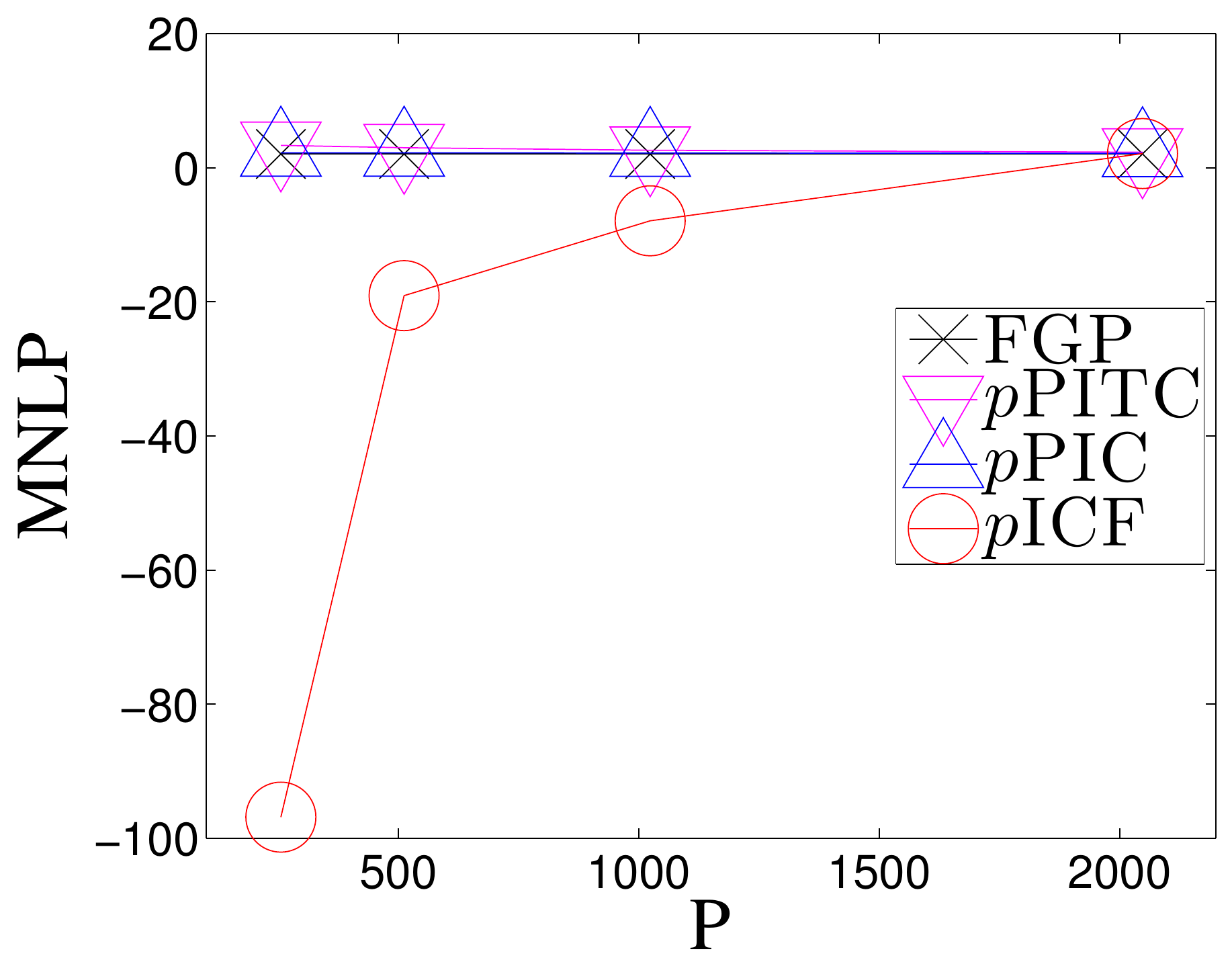} \\
\vspace{-0mm} \hspace{-5mm} (b) AIMPEAK& \hspace{-5mm} (f) SARCOS\\ 
\vspace{-1mm} \hspace{-5mm}
\includegraphics[scale=0.22]{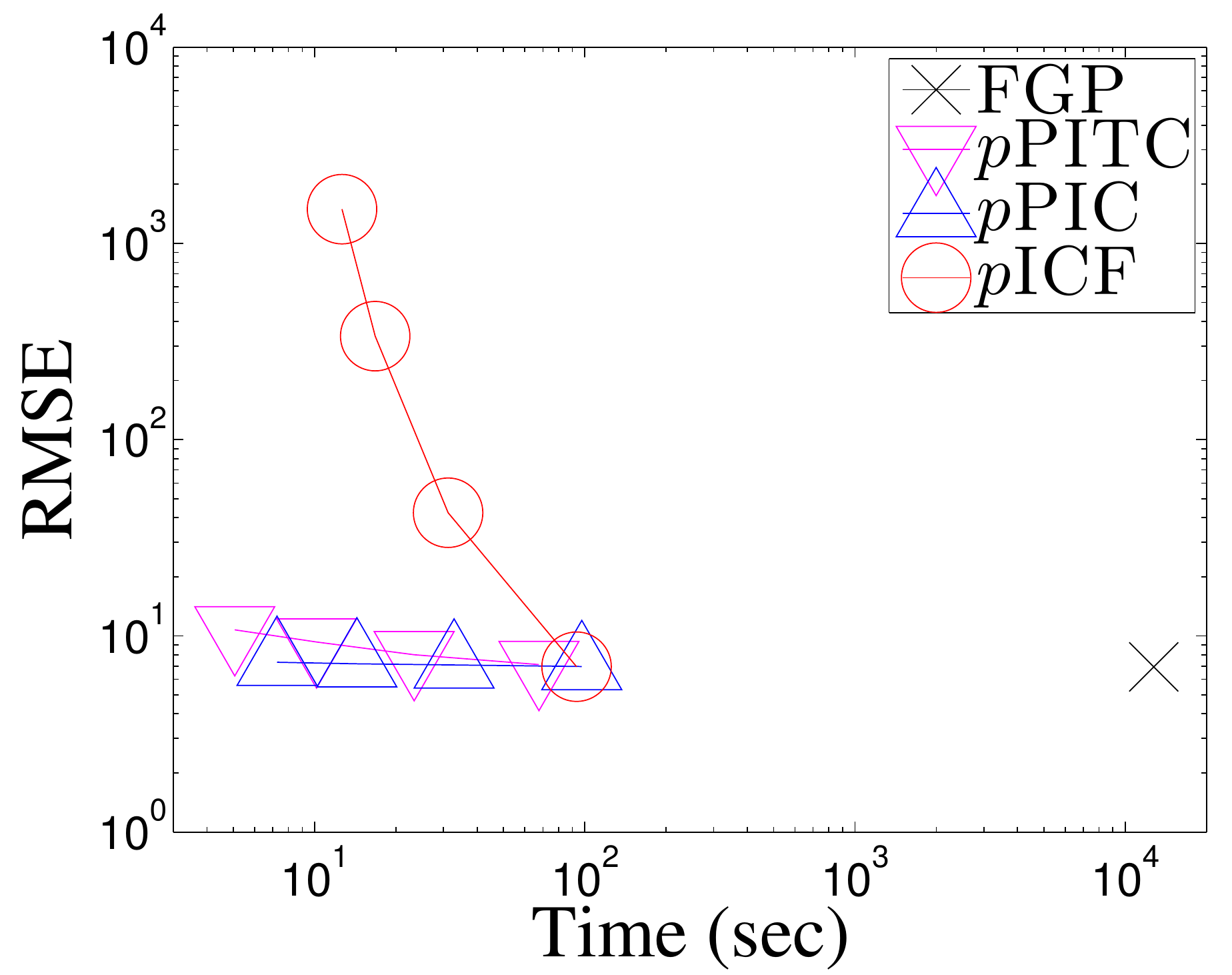} &
\hspace{-5mm}
\includegraphics[scale=0.22]{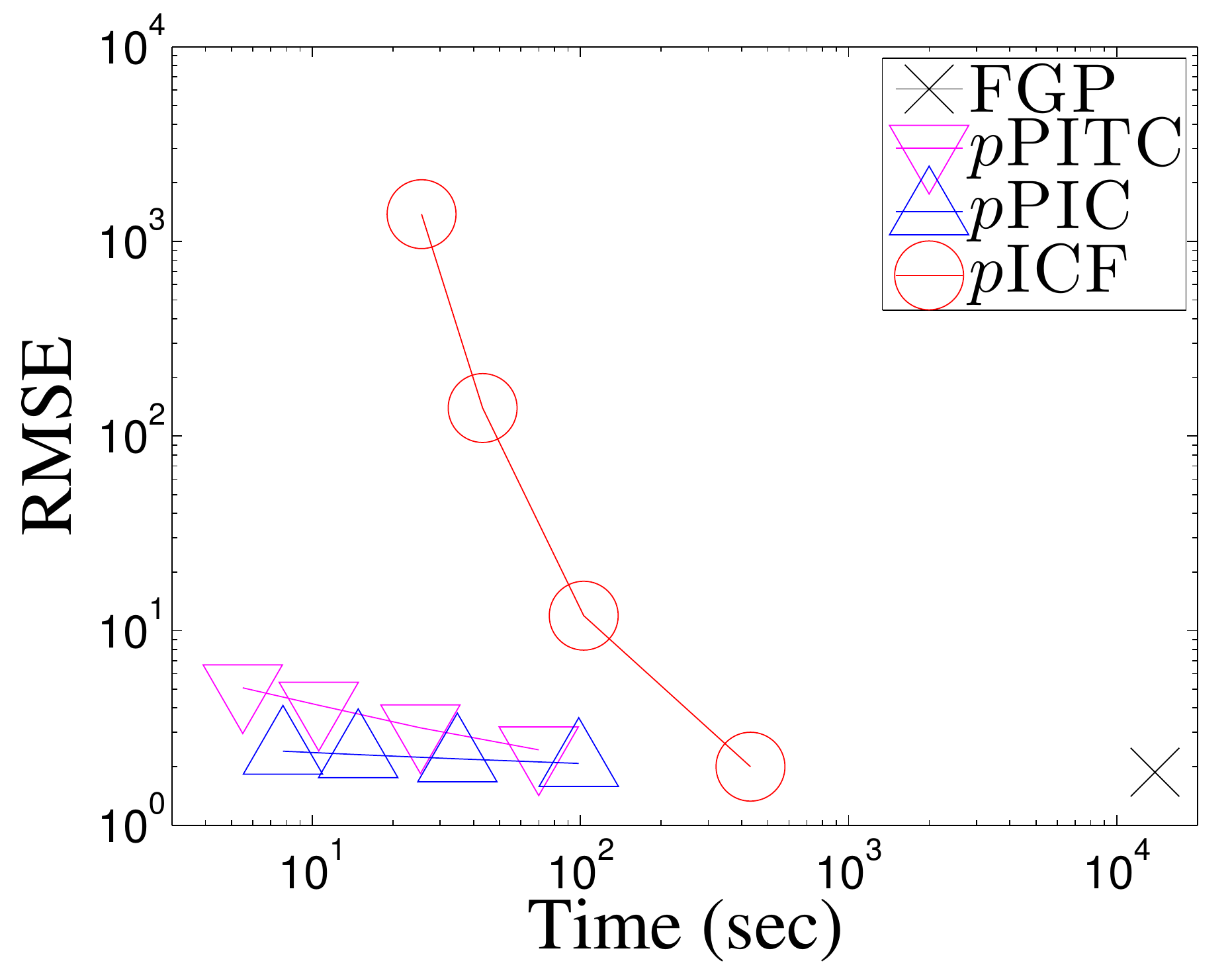} \\
\vspace{-0mm} \hspace{-5mm} (c) AIMPEAK& \hspace{-5mm} (g) SARCOS\\ 
\vspace{-1mm} \hspace{-5mm}
\includegraphics[scale=0.22]{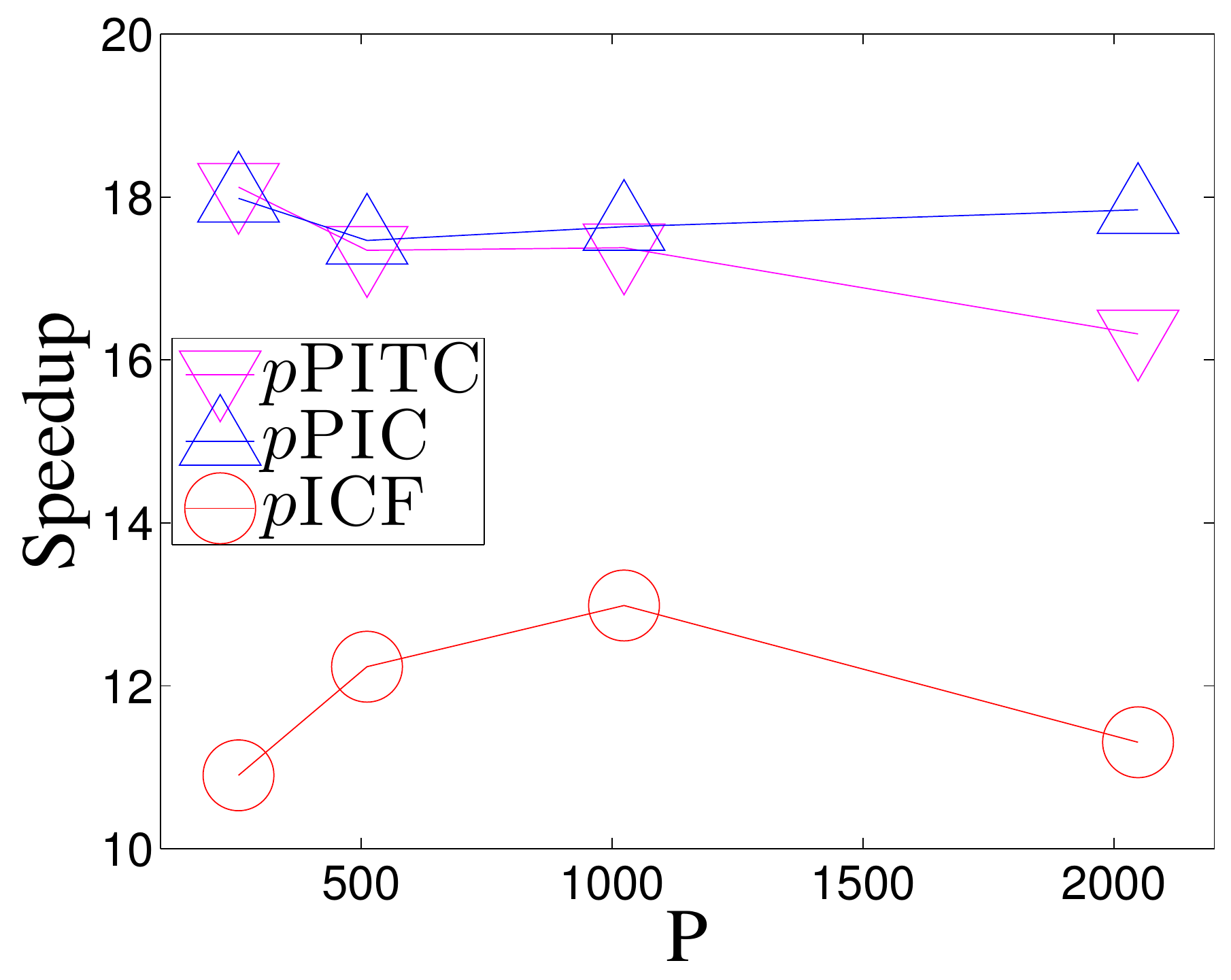} &
\hspace{-5mm}
\includegraphics[scale=0.22]{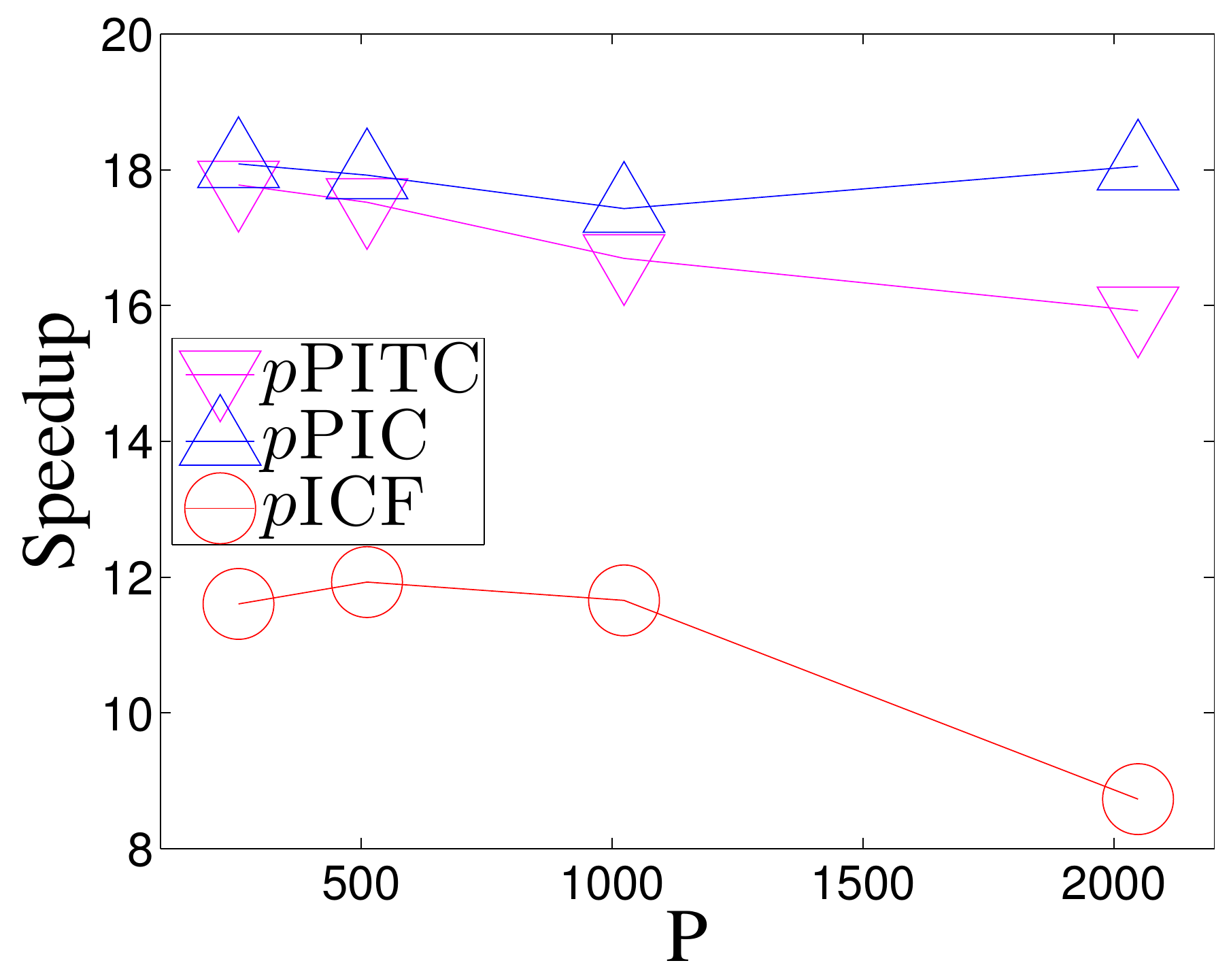} \\
\vspace{-0mm} \hspace{-5mm} (d) AIMPEAK& \hspace{-5mm} (h) SARCOS\\ 
\vspace{-6mm}
%
\end{tabular}
\caption{Performance of parallel GPs with data size $\sset{D}=32000$, number $M=20$ of machines, and varying parameter $P = 256$, $512$, $1024$, $2048$ where $P = \sset{S} = R$ ($P = \sset{S} = R/2$) in the AIMPEAK (SARCOS) domain.\vspace{-4mm}}
\label{fig:varyu} \end{figure}

\vspace{-2mm}
Figs.~\ref{fig:varyu}a and~\ref{fig:varyu}e show that the predictive performance of $p$ICF-based GP is extremely poor when the reduced rank $R$ is not large enough (relative to $\sset{D}$), thus resulting in a poor ICF approximation of the covariance
matrix $\Sigma_{\set{D}\set{D}}$. 
In addition, it can be observed that the reduced rank $R$ of $p$ICF-based GP needs to be much larger than the support set size $\sset{S}$ of $p$PITC and $p$PIC in order to achieve comparable predictive performance.  
These results also indicate that the heuristic $R=\sqrt{\sset{D}}$, which is used by \cite{Chang07} to determine the reduced rank $R$, fails to work well in both our datasets (e.g., $R=1024 > \sqrt{32000}\approx 179$).
    
From Figs.~\ref{fig:varyu}b and~\ref{fig:varyu}f, it can be observed that $p$ICF-based GP incurs negative MNLP for $R\leq 1024$ ($R\leq 2048$) in the AIMPEAK (SARCOS) domain. This is because $p$ICF-based GP cannot guarantee positivity of predictive variance, as explained in Remark $2$ after Theorem~\ref{thm:icfe}. But, it appears that when $R$ is sufficiently large (i.e., $R= 2048$ ($R= 4096$) in the AIMPEAK (SARCOS) domain), this problem can be alleviated.

It can be observed in Figs.~\ref{fig:varyu}c and~\ref{fig:varyu}g that $p$PITC and $p$PIC are significantly more time-efficient than FGP (i.e., $2$-$4$ orders of magnitude faster) while achieving comparable predictive performance.
To ensure high predictive performance, $p$ICF-based GP has to select a large enough rank $R= 2048$ ($R= 4096$) in the AIMPEAK (SARCOS) domain, thus making it less time-efficient than $p$PITC and $p$PIC.
But, it can still incur $1$-$2$ orders of magnitude less time than FGP.
These results indicate that $p$PITC and $p$PIC are more capable than $p$ICF-based GP of meeting the real-time prediction requirement of a time-critical application/system.

Figs.~\ref{fig:varyu}d and~\ref{fig:varyu}h show that $p$PITC and $p$PIC achieve better speedups than $p$ICF-based GP.\vspace{-2mm}
%
\subsubsection{Summary of results}
\vspace{-2mm}
$p$PIC and $p$ICF-based GP are significantly more time-efficient and scalable than FGP (i.e., $1$-$4$ orders of magnitude faster) while achieving comparable predictive performance, hence justifying the practicality of their structural assumptions.
$p$PITC and $p$PIC are expected to be more time-efficient than $p$ICF-based GP with an increasing number $M$ of machines because their incurred time decreases at a faster rate than that of $p$ICF-based GP.
Since the predictive performances of $p$PITC and $p$PIC drop slightly   (i.e., more stable) with smaller $\sset{S}$ as compared to that of $p$ICF-based GP dropping rapidly with smaller $R$,
$p$PITC and $p$PIC are more capable than $p$ICF-based GP of meeting the real-time prediction requirement of time-critical applications.
The speedups of our parallel GPs over their centralized counterparts improve with more data but deviate further from ideal speedup with larger number of machines.\vspace{-2mm}
\section{Conclusion}
\vspace{-2mm}
This paper describes parallel GP regression methods called $p$PIC and $p$ICF-based GP that, respectively, distribute the computational load of the centralized PIC and ICF-based GP among parallel machines to achieve greater time efficiency and scalability. Analytical and empirical results have demonstrated that our parallel GPs are significantly more time-efficient and scalable than their centralized counterparts and FGP while achieving predictive performance comparable to FGP. As a result, by exploiting large clusters of machines, our parallel GPs become substantially more capable of performing real-time predictions necessary in many time-critical applications/systems. We have also implemented $p$PITC and $p$PIC in the MapReduce framework for running in a Linux server with $2$ Intel$\circledR$ Xeon$\circledR$ CPU E$5$-$2670$ at $2.60$~GHz and $96$~GB memory (i.e., $16$ cores); due to shared memory, they incur slightly longer time than that in a cluster of $16$ computing nodes.
We plan to release the source code at http://code.google.com/p/pgpr/.

\noindent{\bf Acknowledgments.} 
This work was supported by Singapore-MIT Alliance Research \& Technology Subaward Agreements No. $28$ R-$252$-$000$-$502$-$592$ \& No. $33$ R-$252$-$000$-$509$-$592$.

\clearpage

\bibliographystyle{natbib}
\bibliography{traffic}

\ifthenelse{\value{sol}=1}{
\clearpage

\appendix
\section{Proof of Theorem~\ref{thm:ppitc}} 
\label{try}
The proof of
Theorem~\ref{thm:ppitc} is previously reported in Appendix A of
\cite{chen12} and
reproduced in this section (i.e., Section~\ref{try}) to reflect our notations.
%
%

We have to first simplify the
$\Gamma_{\set{U}\set{D}}\left(\Gamma_{\set{D}\set{D}}+\Lambda\right)^{-1}$
term in the expressions of $\mu_{\set{U}|\set{D}}^{\mbox{\tiny PITC}}$
(\ref{eq:pitc.mu}) and $\Sigma_{\set{U}\set{U}|D}^{\mbox{\tiny PITC}} $
(\ref{eq:pitc.var}).  
\begin{equation} \hspace{-2mm} \begin{array}{l}
  \left(\Gamma_{\set{D}\set{D}}+\Lambda\right)^{-1}\\ \displaystyle =
  \left(\Sigma_{\set{D}\set{S}}\Sigma_{\set{S}\set{S}}^{-1}\Sigma_{\set{S}\set{D}}+\Lambda\right)^{-1}\\
  \displaystyle = \Lambda^{-1}-\Lambda^{-1}\Sigma_{\set{D}\set{S}}
  \left(\Sigma_{\set{S}\set{S}}+\Sigma_{\set{S}\set{D}}\Lambda^{-1}\Sigma_{\set{D}\set{S}}\right)^{-1}
  \Sigma_{\set{S}\set{D}}\Lambda^{-1}\\ 
\displaystyle
=\Lambda^{-1}-\Lambda^{-1}\Sigma_{\set{D}\set{S}}\ddot{\Sigma}_{\set{S}\set{S}}^{-1}\Sigma_{\set{S}\set{D}}\Lambda^{-1}
\ .  \end{array} \label{eq:pitck} \end{equation} The second equality
follows from matrix inversion lemma.  The last equality is due to
\begin{equation} \begin{array}{l}
  \displaystyle\Sigma_{\set{S}\set{S}}+\Sigma_{\set{S}\set{D}}\Lambda^{-1}\Sigma_{\set{D}\set{S}}\\
  =\displaystyle
  \Sigma_{\set{S}\set{S}}+\sum_{m=1}^{M}\Sigma_{\set{S}\set{D}_m}\Sigma_{\set{D}_m\set{D}_m|\set{S}}^{-1}\Sigma_{\set{D}_m\set{S}}\\
  = \displaystyle
  \Sigma_{\set{S}\set{S}}+\sum_{m=1}^{M}\dot{\Sigma}_{\set{S}\set{S}}^m
  =\displaystyle \ddot{\Sigma}_{\set{S}\set{S}} \ .  \end{array}
  \label{eq:aa} \end{equation} 
Using (\ref{eq:sparseq}) and
  (\ref{eq:pitck}), 
  \begin{equation} 
    \begin{array}{l}
      \Gamma_{{\set{U}_m}\set{D}}\left(\Gamma_{\set{D}\set{D}}+\Lambda\right)^{-1}\\
      =\displaystyle\Sigma_{{\set{U}_m}\set{S}}\Sigma_{\set{S}\set{S}}^{-1}\Sigma_{\set{S}\set{D}}\left(\Lambda^{-1}-\Lambda^{-1}\Sigma_{\set{D}\set{S}}\ddot{\Sigma}_{\set{S}\set{S}}^{-1}\Sigma_{\set{S}\set{D}}\Lambda^{-1}\right)\\
      =\displaystyle\Sigma_{{\set{U}_m}\set{S}}\Sigma_{\set{S}\set{S}}^{-1}\left(\ddot{\Sigma}_{\set{S}\set{S}}-\Sigma_{\set{S}\set{D}}\Lambda^{-1}\Sigma_{\set{D}\set{S}}\right)\ddot{\Sigma}_{\set{S}\set{S}}^{-1}\Sigma_{\set{S}\set{D}}\Lambda^{-1}\\
      =\displaystyle\Sigma_{{\set{U}_m}\set{S}}\ddot{\Sigma}_{\set{S}\set{S}}^{-1}\Sigma_{\set{S}\set{D}}\Lambda^{-1}
    \end{array} \label{eq:pitck2} \end{equation} The third equality is
    due to (\ref{eq:aa}).

For each machine $m= 1,\ldots, M$, we can now prove that 
%
\begin{align*} 
 \mu_{{\set{U}_m}|\set{D}}^{\mbox{\tiny PITC}}
&=\displaystyle\mu_{\set{U}_m}+\Gamma_{{\set{U}_m}\set{D}}\left(\Gamma_{\set{D}\set{D}}+\Lambda\right)^{-1}\left(y_\set{D}-\mu_\set{D} \right)\\
&=\displaystyle \mu_{\set{U}_m}+\Sigma_{{\set{U}_m}\set{S}}\ddot{\Sigma}_{\set{S}\set{S}}^{-1}\Sigma_{\set{S}\set{D}}\Lambda^{-1}\left(y_\set{D}-\mu_\set{D} \right)\\
&=\displaystyle \mu_{\set{U}_m}+\Sigma_{{\set{U}_m}\set{S}}\ddot{\Sigma}_{\set{S}\set{S}}^{-1}\ddot{y}_{\set{S}}\\
&= \widehat{\mu}_{{\set{U}_m}} \ .
  \label{eq:pitc.mur}
\end{align*}
The first equality is by definition (\ref{eq:pitc.mu}).
The second equality is due to (\ref{eq:pitck2}).
The third equality follows from 
$
\Sigma_{\set{S}\set{D}}\Lambda^{-1}\left(y_\set{D}-\mu_\set{D} \right)
=
\sum_{m=1}^{M}\Sigma_{\set{S}\set{D}_m}\Sigma_{\set{D}_m\set{D}_m|\set{S}}^{-1}\left(y_{\set{D}_m}-\mu_{\set{D}_m}
\right) = \sum_{m=1}^{M} \dot{y}^m_{\set{S}}
=\ddot{y}_{\set{S}}$. 
Also,
%
\begin{equation}
  \begin{array}{l} 
    \Sigma_{{\set{U}_m}{\set{U}_m}|\set{D}}^{\mbox{\tiny PITC}}\\
   \displaystyle=\Sigma_{{\set{U}_m}{\set{U}_m}}-\Gamma_{{\set{U}_m}\set{D}}\left(\Gamma_{\set{D}\set{D}}+\Lambda\right)^{-1}\Gamma_{\set{D}{\set{U}_m}}\\
 \displaystyle    =\Sigma_{{\set{U}_m}{\set{U}_m}}-\Sigma_{{\set{U}_m}\set{S}}\ddot{\Sigma}_{\set{S}\set{S}}^{-1}\Sigma_{\set{S}\set{D}}\Lambda^{-1}\Sigma_{\set{D}\set{S}}\Sigma_{\set{S}\set{S}}^{-1}\Sigma_{\set{S}{\set{U}_m}}\\
  \displaystyle    =\Sigma_{{\set{U}_m}{\set{U}_m}}-\Sigma_{{\set{U}_m}\set{S}}\ddot{\Sigma}_{\set{S}\set{S}}^{-1}\left(\ddot{\Sigma}_{\set{S}\set{S}}- \Sigma_{\set{S}\set{S}}\right)
  \Sigma_{\set{S}\set{S}}^{-1}\Sigma_{\set{S}{\set{U}_m}}\\
    \displaystyle =\Sigma_{{\set{U}_m}{\set{U}_m}}-\Sigma_{{\set{U}_m}\set{S}}\left(\Sigma_{\set{S}\set{S}}^{-1}-\ddot{\Sigma}_{\set{S}\set{S}}^{-1}\right)\Sigma_{\set{S}{\set{U}_m}}\\
    = \widehat{\Sigma}_{{\set{U}_m}{\set{U}_m}} \ .
  \end{array} 
      \label{eq:pitc.varr} 
\end{equation}
The first equality is by definition (\ref{eq:pitc.var}).
The second equality follows from (\ref{eq:sparseq}) and
(\ref{eq:pitck2}).  The third equality is due to (\ref{eq:aa}).

Since our primary interest in the work of this paper is to provide the predictive means and their corresponding predictive variances, the above equivalence results suffice.  However, if the entire predictive covariance matrix $\widehat{\Sigma}_{\set{U}\set{U}}$ for any set $\set{U}$ of inputs is desired (say, to calculate the joint entropy), then  
it is necessary to compute $\widehat{\Sigma}_{\set{U}_i\set{U}_j}$ for $i, j =1,\ldots, M$ such that $i\neq j$. Define 
\begin{equation}
\widehat{\Sigma}_{\set{U}_i\set{U}_j}\triangleq \Sigma_{{\set{U}_i}{\set{U}_j}}-\Sigma_{{\set{U}_i}\set{S}}\left(\Sigma_{\set{S}\set{S}}^{-1}-\ddot{\Sigma}_{\set{S}\set{S}}^{-1}\right)\Sigma_{\set{S}{\set{U}_j}}
\label{ohboypitc}
\end{equation}
for $i, j =1,\ldots, M$ such that $i\neq j$. So, for a machine $i$ to compute $\widehat{\Sigma}_{\set{U}_i\set{U}_j}$, it has to receive $\set{U}_j$ from machine $j$.

Similar to (\ref{eq:pitc.varr}), we can prove the equivalence result $\widehat{\Sigma}_{\set{U}_i\set{U}_j} = \Sigma_{\set{U}_i\set{U}_j|\set{D}}^{\mbox{\tiny{PITC}}}$ for any two machines $i, j =1,\ldots, M$ such that $i\neq j$.

%
\section{Proof of Theorem~\ref{thm:ppic}}
We will first derive the expressions of four components useful for
completing the proof later. For each machine $m=1,\ldots, M$,
\vspace{-0mm}\begin{equation}
\hspace{-1.8mm}
  \begin{array}{l}
\displaystyle
\widetilde{\Gamma}_{\set{U}_m\set{D}}\Lambda^{-1} (y_\set{D}-\mu_\set{D})\\
\displaystyle
= \sum_{i\neq
m}\Gamma_{\set{U}_m\set{D}_i}\Sigma_{\set{D}_i\set{D}_i|\set{S}}^{-1}(y_{\set{D}_i}-\mu_{\set{D}_i})\\
\quad+\;\Sigma_{\set{U}_m\set{D}_m}\Sigma_{\set{D}_m\set{D}_m|\set{S}}^{-1}(y_{\set{D}_m}-\mu_{\set{D}_m})\\ 
\displaystyle
= \Sigma_{\set{U}_m\set{S}}\Sigma_{\set{S}\set{S}}^{-1}\sum_{i\neq
m}\left(\Sigma_{\set{S}\set{D}_i}
\Sigma_{\set{D}_i\set{D}_i|\set{S}}^{-1}(y_{\set{D}_i}-\mu_{\set{D}_i})\right) + \dot{y}_{\set{U}_m}^m \\
\displaystyle
= \Sigma_{\set{U}_m\set{S}}\Sigma_{\set{S}\set{S}}^{-1}\sum_{i\neq m}\dot{y}_{\set{S}}^i
+ \dot{y}_{\set{U}_m}^m \\
\displaystyle
= \Sigma_{\set{U}_m\set{S}}\Sigma_{\set{S}\set{S}}^{-1}(\ddot{y}_{\set{S}}-\dot{y}_{\set{S}}^m)
+ \dot{y}_{\set{U}_m}^m \ .
\end{array}
\label{pf:p1}
\end{equation}
The first two equalities expand the first component using the definition
of $\Lambda$ (Theorem~\ref{thm:ppitc}), (\ref{eq:lsf}), (\ref{eq:sparseq}), (\ref{eq:pick}), and (\ref{eq:pickernel}). The last two equalities exploit (\ref{eq:lsf}) and
(\ref{eq:gsf}). 
\vspace{-0mm}\begin{equation}
\hspace{-1.7mm}
\begin{array}{l}
\displaystyle
\widetilde{\Gamma}_{\set{U}_m\set{D}}\Lambda^{-1} \Sigma_{\set{D}\set{S}}\\
\displaystyle
= \sum_{i\neq m}\Gamma_{\set{U}_m\set{D}_i}\Sigma_{\set{D}_i\set{D}_i|\set{S}}^{-1}\Sigma_{\set{D}_iS} 
+\Sigma_{\set{U}_m\set{D}_m}\Sigma_{\set{D}_m\set{D}_m|\set{S}}^{-1}\Sigma_{\set{D}_m\set{S}} \\
\displaystyle
= \Sigma_{\set{U}_m\set{S}}\Sigma_{\set{S}\set{S}}^{-1}\sum_{i\neq m}\left(\Sigma_{\set{S}\set{D}_i}
\Sigma_{\set{D}_i\set{D}_i|\set{S}}^{-1}\Sigma_{\set{D}_iS}\right)\\
\quad+\;\Sigma_{\set{U}_m\set{D}_m}\Sigma_{\set{D}_m\set{D}_m|\set{S}}^{-1}\Sigma_{\set{D}_m\set{S}} \\
%
%
\displaystyle
= \Sigma_{\set{U}_m\set{S}}\Sigma_{\set{S}\set{S}}^{-1}\sum_{i\neq m}\dot{\Sigma}_{\set{S}\set{S}}^i
+\dot{\Sigma}_{\set{U}_m\set{S}}^m \\
%
\displaystyle
= \Sigma_{\set{U}_m\set{S}}\Sigma_{\set{S}\set{S}}^{-1}\left(\ddot{\Sigma}_{\set{S}\set{S}}
    -\dot{\Sigma}_{\set{S}\set{S}}^m-{\Sigma}_{\set{S}\set{S}}\right)
+\dot{\Sigma}_{\set{U}_m\set{S}}^m \\
%
%
\displaystyle
=\Sigma_{\set{U}_m\set{S}}\Sigma_{\set{S}\set{S}}^{-1}\ddot{\Sigma}_{\set{S}\set{S}}-{\Phi}_{\set{U}_m\set{S}}^m\ .
\end{array}
\label{pf:p2}
\end{equation}
The first two equalities expand the second component by the same trick as that in
(\ref{pf:p1}). 
The third and fourth equalities exploit (\ref{eq:lsc}) and (\ref{eq:gsc}), respectively. The last equality is due to (\ref{eq:keys}).

Let $\alpha_{\set{U_m}\set{S}}\triangleq\Sigma_{\set{U_m}\set{S}}\Sigma_{\set{S}\set{S}}^{-1}$ and its transpose
is $\alpha_{\set{S}\set{U_m}}$. By using similar tricks in (\ref{pf:p1}) and
(\ref{pf:p2}), we can derive the expressions of the remaining two components.

%
\vspace{-0mm}\begin{equation}
\hspace{-1.8mm}
\begin{array}{l}
\displaystyle
\widetilde{\Gamma}_{\set{U}_m\set{D}}\Lambda^{-1}\widetilde{\Gamma}_{\set{D}\set{U}_m}\\
%
\displaystyle
= \sum_{i\neq m}\Gamma_{\set{U}_m\set{D}_i}\Sigma_{\set{D}_i\set{D}_i|\set{S}}^{-1}\Gamma_{\set{D}_i\set{U}_m} 
+\Sigma_{\set{U}_m\set{D}_m}\Sigma_{\set{D}_m\set{D}_m|\set{S}}^{-1}\Sigma_{\set{D}_m\set{U}_m} \\
\displaystyle
= \Sigma_{\set{U}_m\set{S}}\Sigma_{\set{S}\set{S}}^{-1}\sum_{i\neq m}\left(\Sigma_{\set{S}\set{D}_i}
\Sigma_{\set{D}_i\set{D}_i|\set{S}}^{-1}\Sigma_{\set{D}_iS}\right)\Sigma_{\set{S}\set{S}}^{-1}\Sigma_{\set{S}\set{U}_m}+\dot{\Sigma}_{\set{U}_m\set{U}_m}^m \\
\displaystyle
= \alpha_{\set{U}_m\set{S}}\sum_{i\neq m}\left(
    \dot{\Sigma}_{\set{S}\set{S}}^i
\right)\alpha_{\set{S}\set{U}_m}
+ \dot{\Sigma}_{\set{U}_m\set{U}_m}^m \\
\displaystyle
= \alpha_{\set{U}_m\set{S}}\left(\ddot{\Sigma}_{\set{S}\set{S}}-\dot{\Sigma}_{\set{S}\set{S}}^m-{\Sigma}_{\set{S}\set{S}}
\right)\alpha_{\set{S}\set{U}_m} + \dot{\Sigma}_{\set{U}_m\set{U}_m}^m \\
\displaystyle
=\alpha_{\set{U}_m\set{S}}\ddot{\Sigma}_{\set{S}\set{S}}\alpha_{\set{S}\set{U}_m}
    -\alpha_{\set{U}_m\set{S}}{\Phi}_{\set{S}\set{U}_m}^m - \alpha_{\set{U}_m\set{S}}\dot{\Sigma}^m_{\set{S}\set{U}_m} 
+ \dot{\Sigma}_{\set{U}_m\set{U}_m}^m \ .
\end{array}
\label{pf:p3}
\end{equation}

For any two machines $i, j =1,\ldots, M$ such that $i\neq j$,
\vspace{-0mm}\begin{equation}
\hspace{-1.8mm}
\begin{array}{l}
\displaystyle
\widetilde{\Gamma}_{\set{U}_i\set{D}}\Lambda^{-1}\widetilde{\Gamma}_{\set{D}\set{U}_j}\\
\displaystyle
=\sum_{m\neq i,j}\Gamma_{\set{U}_i\set{D}_m}\Sigma_{\set{D}_m\set{D}_m|\set{S}}^{-1}\Gamma_{\set{D}_m\set{U}_j}\\ 
\quad +\ \Sigma_{\set{U}_i\set{D}_i}\Sigma_{\set{D}_i\set{D}_i|\set{S}}^{-1}\Gamma_{\set{D}_i\set{U}_j}
 +\Gamma_{\set{U}_i\set{D}_j}\Sigma_{\set{D}_j\set{D}_j|\set{S}}^{-1}\Sigma_{\set{D}_j\set{U}_j}\\ 
\displaystyle
=\Sigma_{\set{U}_i\set{S}}\Sigma_{\set{S}\set{S}}^{-1}\sum_{m\neq
i,j}\left(\Sigma_{\set{S}\set{D}_m}\Sigma_{\set{D}_m\set{D}_m|\set{S}}^{-1}\Sigma_{\set{D}_m\set{S}}\right)
\Sigma_{\set{S}\set{S}}^{-1}\Sigma_{\set{S}\set{U}_j}\\ 
\quad +\
\Sigma_{\set{U}_i\set{D}_i}\Sigma_{\set{D}_i\set{D}_i|\set{S}}^{-1}\Sigma_{\set{D}_iS}\Sigma_{\set{S}\set{S}}^{-1}\Sigma_{\set{S}\set{U}_j}\\
\quad +\ \Sigma_{\set{U}_i\set{S}}\Sigma_{\set{S}\set{S}}^{-1}\Sigma_{\set{S}\set{D}_j}\Sigma_{\set{D}_j\set{D}_j|\set{S}}^{-1}\Sigma_{\set{D}_j\set{U}_j}\\ 
%
%
\displaystyle
=\alpha_{\set{U}_i\set{S}}\left(\ddot{\Sigma}_{\set{S}\set{S}}-\dot{\Sigma}_{\set{S}\set{S}}^i
  -\dot{\Sigma}_{\set{S}\set{S}}^j-{\Sigma}_{\set{S}\set{S}}\right)\alpha_{\set{S}\set{U}_j}\\
  \quad +\ \dot{\Sigma}^i_{\set{U}_i\set{S}}\alpha_{\set{S}\set{U}_j}
 +{\alpha}_{\set{U}_i\set{S}}\dot{\Sigma}^j_{\set{S}\set{U}_j}\\ 
\displaystyle
=\alpha_{\set{U}_i\set{S}}\left(\ddot{\Sigma}_{\set{S}\set{S}}+{\Sigma}_{\set{S}\set{S}}\right)\alpha_{\set{S}\set{U}_j} 
 -\alpha_{\set{U}_i\set{S}}\left(\dot{\Sigma}_{\set{S}\set{S}}^i+{\Sigma}_{\set{S}\set{S}}\right)\alpha_{\set{S}\set{U}_j}\\
\quad-\ \alpha_{\set{U}_i\set{S}}\left(\dot{\Sigma}_{\set{S}\set{S}}^j+{\Sigma}_{\set{S}\set{S}}\right)\alpha_{\set{S}\set{U}_j} 
+ \dot{\Sigma}^i_{\set{U}_i\set{S}}\alpha_{\set{S}\set{U}_j}
 +{\alpha}_{\set{U}_i\set{S}}\dot{\Sigma}^j_{\set{S}\set{U}_j}\\ 
%
%
\displaystyle
= \alpha_{\set{U}_i\set{S}}\left(\ddot{\Sigma}_{\set{S}\set{S}}+{\Sigma}_{\set{S}\set{S}}\right)\alpha_{\set{S}\set{U}_j}
-
\left(\alpha_{\set{U}_i\set{S}}\dot{\Sigma}_{\set{S}\set{S}}^i+\alpha_{\set{U}_i\set{S}}{\Sigma}_{\set{S}\set{S}}\right.\\
\left. \quad -\ \dot{\Sigma}^i_{\set{U}_i\set{S}}\right)\alpha_{\set{S}\set{U}_j}
- \alpha_{\set{U}_i\set{S}}\left(\dot{\Sigma}_{\set{S}\set{S}}^j \alpha_{\set{S}\set{U}_j}+{\Sigma}_{\set{S}\set{S}}\alpha_{\set{S}\set{U}_j}-\dot{\Sigma}^j_{\set{S}\set{U}_j}\right)\\ 
\displaystyle
  =\alpha_{\set{U}_i\set{S}}\left(\ddot{\Sigma}_{\set{S}\set{S}}+{\Sigma}_{\set{S}\set{S}}\right)\alpha_{\set{S}\set{U}_j}
-{\Phi}_{\set{U}_i\set{S}}^i \alpha_{\set{S}\set{U}_j} -{\alpha}_{\set{U}_i\set{S}} {\Phi}_{\set{S}\set{U}_j}^j\\
\displaystyle
  =\alpha_{\set{U}_i\set{S}}\ddot{\Sigma}_{\set{S}\set{S}}\alpha_{\set{S}\set{U}_j}
  +\Sigma_{\set{U}_i\set{S}}{\Sigma}_{\set{S}\set{S}}^{-1}\Sigma_{\set{S}\set{U}_j}
-{\Phi}_{\set{U}_i\set{S}}^i \alpha_{\set{S}\set{U}_j} -{\alpha}_{\set{U}_i\set{S}} {\Phi}_{\set{S}\set{U}_j}^j\\
  \end{array}
  \label{pf:p4}
\end{equation}

For each machine $m=1,\ldots, M$, we can now prove that
$$
\begin{array}{l}
\mu^{\mbox{\tiny{PIC}}}_{\set{U}_m|\set{D}}\\
= \mu_{\set{U}_m} +
\widetilde{\Gamma}_{\set{U}_m\set{D}}\left(\Gamma_{\set{D}\set{D}}+\Lambda\right)^{-1}(y_\set{D}-\mu_\set{D})\\
=\mu_{\set{U}_m} + \widetilde{\Gamma}_{\set{U}_m\set{D}}\Lambda^{-1}(y_\set{D}-\mu_\set{D})\\
\quad -\ \widetilde{\Gamma}_{\set{U}_m\set{D}}\Lambda^{-1}\Sigma_{\set{D}\set{S}}\ddot{\Sigma}_{\set{S}\set{S}}^{-1}\Sigma_{\set{S}\set{D}}\Lambda^{-1}(y_\set{D}-\mu_\set{D})\\
=\mu_{\set{U}_m} + \widetilde{\Gamma}_{\set{U}_m\set{D}}\Lambda^{-1}(y_\set{D}-\mu_\set{D})
- \widetilde{\Gamma}_{\set{U}_m\set{D}}\Lambda^{-1}\Sigma_{\set{D}\set{S}}\ddot{\Sigma}_{\set{S}\set{S}}^{-1}\ddot{y}_{\set{S}}\\
=\mu_{\set{U}_m} + \Sigma_{\set{U}_m\set{S}}\Sigma_{\set{S}\set{S}}^{-1}(\ddot{y}_{\set{S}}-\dot{y}_{\set{S}}^m)
+\dot{y}_{\set{U}_m}^m\\
\quad -\ \widetilde{\Gamma}_{\set{U}_m\set{D}}\Lambda^{-1}\Sigma_{\set{D}\set{S}}\ddot{\Sigma}_{\set{S}\set{S}}^{-1}\ddot{y}_{\set{S}}\\
=\mu_{\set{U}_m} + \Sigma_{\set{U}_m\set{S}}\Sigma_{\set{S}\set{S}}^{-1}(\ddot{y}_{\set{S}}-\dot{y}_{\set{S}}^m)
+\dot{y}_{\set{U}_m}^m \\
\quad-\left(\Sigma_{\set{U}_m\set{S}}\Sigma_{\set{S}\set{S}}^{-1}\ddot{\Sigma}_{\set{S}\set{S}}-{\Phi}_{\set{U}_m\set{S}}^m
\right) \ddot{\Sigma}_{\set{S}\set{S}}^{-1}\ddot{y}_{\set{S}}\\
=\mu_{\set{U}_m} +\left({\Phi}_{\set{U}_m\set{S}}^m\ddot{\Sigma}_{\set{S}\set{S}}^{-1}\ddot{y}_{\set{S}}
-\Sigma_{\set{U}_m\set{S}}\Sigma_{\set{S}\set{S}}^{-1}\dot{y}_{\set{S}}^m\right) +\dot{y}_{\set{U}_m}^m\\
=  \widehat{\mu}_{\set{U}_m}^+\ .  
\end{array}
$$
The first equality is by definition (\ref{eq:picmu}).
The second equality is due to (\ref{eq:pitck}). The third equality is
due to the definition of global summary (\ref{eq:gsf}). The fourth and fifth equalities
are due to (\ref{pf:p1}) and (\ref{pf:p2}), respectively. Also,
%

$$
\begin{array}{l}
\Sigma_{\set{U}_m\set{U}_m|\set{D}}^{\mbox{\tiny{PIC}}}\\
= \Sigma_{\set{U}_m\set{U}_m}- 
\widetilde{\Gamma}_{\set{U}_m\set{D}}\left(\Gamma_{\set{D}\set{D}}+\Lambda\right)^{-1}\widetilde{\Gamma}_{\set{D}\set{U}_m}\\
= \Sigma_{\set{U}_m\set{U}_m}-\widetilde{\Gamma}_{\set{U}_m\set{D}}\Lambda^{-1}\widetilde{\Gamma}_{\set{D}\set{U}_m}
+\widetilde{\Gamma}_{\set{U}_m\set{D}}\Lambda^{-1}\Sigma_{\set{D}\set{S}}\ddot{\Sigma}_{\set{S}\set{S}}^{-1}
\Sigma_{\set{S}\set{D}}\Lambda^{-1}\widetilde{\Gamma}_{\set{D}\set{U}_m}\\
=\Sigma_{\set{U}_m\set{U}_m}
-\widetilde{\Gamma}_{\set{U}_m\set{D}}\Lambda^{-1}\widetilde{\Gamma}_{\set{D}\set{U}_m}\\
\quad+\left(\alpha_{\set{U}_m\set{S}}\ddot{\Sigma}_{\set{S}\set{S}}-{\Phi}_{\set{U}_m\set{S}}^m\right)
\ddot{\Sigma}_{\set{S}\set{S}}^{-1}
\left(\ddot{\Sigma}_{\set{S}\set{S}}\alpha_{\set{S}\set{U}_m}-{\Phi}_{\set{S}\set{U}_m}^m\right)\\
%
=\Sigma_{\set{U}_m\set{U}_m} -
\alpha_{\set{U}_m\set{S}}\ddot{\Sigma}_{\set{S}\set{S}}\alpha_{\set{S}\set{U}_m}
    +\alpha_{\set{U}_m\set{S}}{\Phi}_{\set{S}\set{U}_m}^m
    +\alpha_{\set{U}_m\set{S}}\dot{\Sigma}^m_{\set{S}\set{U}_m}\\ 
    \quad -\ \dot{\Sigma}_{\set{U}_m\set{U}_m}^m +
\alpha_{\set{U}_m\set{S}}\ddot{\Sigma}_{\set{S}\set{S}}\alpha_{\set{S}\set{U}_m}
    -\alpha_{\set{U}_m\set{S}}\Phi_{\set{S}\set{U}_m}^m
    -\Phi_{\set{U}_m\set{S}}^m\alpha_{\set{S}\set{U}_m}\\ 
    \quad +\ {\Phi}_{\set{U}_m\set{S}}^m\ddot{\Sigma}_{\set{S}\set{S}}^{-1}{\Phi}_{\set{S}\set{U}_m}^m \\
%
%
=\Sigma_{\set{U}_m\set{U}_m}
-\left(\Phi_{\set{U}_m\set{S}}^m\alpha_{\set{S}\set{U}_m}
-\alpha_{\set{U}_m\set{S}}\dot{\Sigma}^m_{\set{S}\set{U}_m}
-{\Phi}_{\set{U}_m\set{S}}^m\ddot{\Sigma}_{\set{S}\set{S}}^{-1}{\Phi}_{\set{S}\set{U}_m}^m\right)\\
\quad -\ \dot{\Sigma}_{\set{U}_m\set{U}_m}^m\\
= \widehat{\Sigma}_{\set{U}_m\set{U}_m}^+\ .
%
%
\end{array}
$$
The first equality is by definition (\ref{eq:picvar}).
The second equality is due to (\ref{eq:pitck}). The third equality is
due to (\ref{pf:p2}). The fourth equality is due to (\ref{pf:p3}). The
last two equalities are by definition (\ref{eq:lacov}).
%

Since our primary interest in the work of this paper is to provide the predictive means and their corresponding predictive variances, the above equivalence results suffice.  However, if the entire predictive covariance matrix $\widehat{\Sigma}_{\set{U}\set{U}}^+$ for any set $\set{U}$ of inputs is desired (say, to calculate the joint entropy), then  
it is necessary to compute $\widehat{\Sigma}^+_{\set{U}_i\set{U}_j}$ for $i, j =1,\ldots, M$ such that $i\neq j$. Define 
\begin{equation}
\widehat{\Sigma}^+_{\set{U}_i\set{U}_j}\triangleq \Sigma_{\set{U}_i\set{U}_j|\set{S}}+{\Phi}_{\set{U}_i\set{S}}^i\ddot{\Sigma}_{\set{S}\set{S}}^{-1}{\Phi}_{\set{S}\set{U}_j}^j
\label{ohboy}
\end{equation}
for $i, j =1,\ldots, M$ such that $i\neq j$. So, for a machine $i$ to compute $\widehat{\Sigma}^+_{\set{U}_i\set{U}_j}$, it has to receive $\set{U}_j$ and ${\Phi}_{\set{S}\set{U}_j}^j$ from machine $j$.

We can now prove the equivalence result $\widehat{\Sigma}^+_{\set{U}_i\set{U}_j} = \Sigma_{\set{U}_i\set{U}_j|\set{D}}^{\mbox{\tiny{PIC}}}$ for any two machines $i, j =1,\ldots, M$ such that $i\neq j$:
$$
\hspace{-1.8mm}
\begin{array}{l}
\Sigma_{\set{U}_i\set{U}_j|\set{D}}^{\mbox{\tiny{PIC}}}\\
=\Sigma_{\set{U}_i\set{U}_j}
-\widetilde{\Gamma}_{\set{U}_i\set{D}}\Lambda^{-1}\widetilde{\Gamma}_{\set{D}\set{U}_j} +\ \alpha_{\set{U}_i\set{S}}\ddot{\Sigma}_{\set{S}\set{S}}\alpha_{\set{S}\set{U}_j}\\
\quad -\alpha_{\set{U}_i\set{S}}\Phi_{\set{S}\set{U}_j}^j     -\Phi_{\set{U}_i\set{S}}^i\alpha_{\set{S}\set{U}_j}
    +{\Phi}_{\set{U}_i\set{S}}^i\ddot{\Sigma}_{\set{S}\set{S}}^{-1}{\Phi}_{\set{S}\set{U}_j}^j \\
=\Sigma_{\set{U}_i\set{U}_j}-\left(\alpha_{\set{U}_i\set{S}}\ddot{\Sigma}_{\set{S}\set{S}}\alpha_{\set{S}\set{U}_j}
+\Sigma_{\set{U}_i\set{S}}{\Sigma}_{\set{S}\set{S}}^{-1}\Sigma_{\set{S}\set{U}_j}
-{\Phi}_{\set{U}_i\set{S}}^i \alpha_{\set{S}\set{U}_j}\right.\\ 
\left.\quad -\ {\alpha}_{\set{U}_i\set{S}} {\Phi}_{\set{S}\set{U}_j}^j
  \right)
+\alpha_{\set{U}_i\set{S}}\ddot{\Sigma}_{\set{S}\set{S}}\alpha_{\set{S}\set{U}_j}
-\alpha_{\set{U}_i\set{S}}\Phi_{\set{S}\set{U}_j}^j-\Phi_{\set{U}_i\set{S}}^i\alpha_{\set{S}\set{U}_j}\\
\quad+\ {\Phi}_{\set{U}_i\set{S}}^i\ddot{\Sigma}_{\set{S}\set{S}}^{-1}{\Phi}_{\set{S}\set{U}_j}^j\\
=\Sigma_{\set{U}_i\set{U}_j} -\Sigma_{\set{U}_i\set{S}}{\Sigma}_{\set{S}\set{S}}^{-1}\Sigma_{\set{S}\set{U}_j}
+{\Phi}_{\set{U}_i\set{S}}^i\ddot{\Sigma}_{\set{S}\set{S}}^{-1}{\Phi}_{\set{S}\set{U}_j}^j\\
=\Sigma_{\set{U}_i\set{U}_j|\set{S}}+{\Phi}_{\set{U}_i\set{S}}^i\ddot{\Sigma}_{\set{S}\set{S}}^{-1}{\Phi}_{\set{S}\set{U}_j}^j\\
= \widehat{\Sigma}^+_{\set{U}_i\set{U}_j}\ .
\end{array}
$$
The first equality is obtained using a similar trick as the previous
derivation.
The second equality is due to (\ref{pf:p4}). The second last
equality is by the definition of posterior covariance in GP model (\ref{pcov}).
The last equality is by definition (\ref{ohboy}).
\section{Proof of Theorem~\ref{thm:icfe}}
$$
\begin{array}{l}
\mu^{\mbox{\tiny ICF}}_{\set{U}|\set{D}}\\
\displaystyle= \mu_\set{U} + \Sigma_{\set{U}\set{D}}(F^\top F + \sigma^2_n I)^{-1}( y_\set{D} - \mu_\set{D} )\\
\displaystyle= \mu_\set{U} + \Sigma_{\set{U}\set{D}} \Big(\sigma^{-2}_n ( y_\set{D} - \mu_\set{D} )\\
\quad\displaystyle - \sigma^{-4}_n F^\top (I + \sigma^{-2}_n F F^\top)^{-1} F ( y_\set{D} - \mu_\set{D}) \Big)\\
\displaystyle= \mu_\set{U} + \Sigma_{\set{U}\set{D}} \left(\sigma^{-2}_n ( y_\set{D} - \mu_\set{D} ) - \sigma^{-4}_n F^\top \Phi^{-1} \sum^M_{m=1} \dot{y}_m \right)\\
\displaystyle=\mu_\set{U} + \Sigma_{\set{U}\set{D}} \left(\sigma^{-2}_n ( y_\set{D} - \mu_\set{D} ) - \sigma^{-4}_n F^\top \ddot{y} \right)\\
\displaystyle=\mu_\set{U} + \sum^M_{m=1}\Sigma_{\set{U}\set{D_m}}\left(\sigma^{-2}_n (y_{\set{D_m}}-\mu_{\set{D_m}}) - \sigma^{-4}_n F^\top_m \ddot{y}\right)\\
\displaystyle=\mu_\set{U} + \sum^M_{m=1}
\sigma^{-2}_n \Sigma_{\set{U}\set{D_m}}(y_{\set{D_m}}-\mu_{\set{D_m}}) - \sigma^{-4}_n \dot{\Sigma}^\top_m \ddot{y}\\
\displaystyle=\mu_\set{U} + \sum^M_{m=1} \widetilde{\mu}^m_{\set{U}}\\
=\widetilde{\mu}_{\set{U}}\ .
\end{array}
$$
The first equality is by definition (\ref{imu}).
The second equality is due to matrix inversion lemma.
The third equality follows from $I + \sigma^{-2}_n F F^\top = I+\sigma^{-2}_n\sum^M_{m=1} F_m F^\top_m = I+\sigma^{-2}_n\sum^M_{m=1}\Phi_m = \Phi$ and
$F ( y_\set{D} - \mu_\set{D} ) = \sum^M_{m=1} F_m (y_{\set{D_m}}-\mu_{\set{D_m}}) = \sum^M_{m=1} \dot{y}_m$.
The fourth equality is due to (\ref{eq:gsimu}).
The second last equality follows from (\ref{eq:pcimu}).
The last equality is by definition (\ref{eq:icfmu}).

Similarly,
$$
\begin{array}{l}
\Sigma^{\mbox{\tiny ICF}}_{\set{U}\set{U}|\set{D}}\\
\displaystyle= \Sigma_{\set{U}\set{U}}-\Sigma_{\set{U}\set{D}}(F^\top F + \sigma^2_n I)^{-1}\Sigma_{\set{D}\set{U}}\\
\displaystyle= \Sigma_{\set{U}\set{U}}-\Sigma_{\set{U}\set{D}} \Big(\sigma^{-2}_n \Sigma_{\set{D}\set{U}}\\
\quad\displaystyle - \sigma^{-4}_n F^\top (I + \sigma^{-2}_n F F^\top)^{-1} F \Sigma_{\set{D}\set{U}} \Big)\\
\displaystyle= \Sigma_{\set{U}\set{U}}- \Sigma_{\set{U}\set{D}} \left(\sigma^{-2}_n \Sigma_{\set{D}\set{U}} - \sigma^{-4}_n F^\top \Phi^{-1} \sum^M_{m=1} \dot{\Sigma}_m \right)\\
\displaystyle=\Sigma_{\set{U}\set{U}}- \Sigma_{\set{U}\set{D}} \left(\sigma^{-2}_n \Sigma_{\set{D}\set{U}} - \sigma^{-4}_n F^\top \ddot{\Sigma} \right)\\
\displaystyle=\Sigma_{\set{U}\set{U}}- \sum^M_{m=1}\Sigma_{\set{U}\set{D_m}}\left(\sigma^{-2}_n \Sigma_{\set{D}_m\set{U}} - \sigma^{-4}_n F^\top_m \ddot{\Sigma}\right)\\
\displaystyle=\Sigma_{\set{U}\set{U}}- \sum^M_{m=1}\sigma^{-2}_n \Sigma_{\set{U}\set{D_m}}\Sigma_{\set{D_m}\set{U}} - \sigma^{-4}_n \dot{\Sigma}^\top_m \ddot{\Sigma}\\
\displaystyle=\Sigma_{\set{U}\set{U}}- \sum^M_{m=1} \widetilde{\Sigma}^m_{\set{U}\set{U}}\\
=\widetilde{\Sigma}_{\set{U}\set{U}}\ .
\end{array}
$$
The first equality is by definition (\ref{icov}).
The second equality is due to matrix inversion lemma.
The third equality follows from $I + \sigma^{-2}_n F F^\top = \Phi$ and
$F \Sigma_{\set{D}\set{U}} = \sum^M_{m=1} F_m \Sigma_{\set{D}_m \set{U}} = \sum^M_{m=1} \dot{\Sigma}_m$.
The fourth equality is due to (\ref{eq:gsivar}).
The second last equality follows from (\ref{eq:pcivar}).
The last equality is by definition (\ref{eq:icfvar}).
}{}
\end{document}